\documentclass[lettersize,journal]{IEEEtran}
\usepackage{amsmath,amsfonts}
\usepackage{algorithmic}
\usepackage{algorithm}
\usepackage{array}
\usepackage[caption=false,font=normalsize,labelfont=sf,textfont=sf]{subfig}
\usepackage{textcomp}
\usepackage{stfloats}
\usepackage{url}
\usepackage{verbatim}
\usepackage{graphicx}
\usepackage{cite}
\usepackage{wrapfig}
\usepackage{graphicx}

\usepackage{enumitem}
\usepackage{booktabs}
\usepackage{multirow}
\usepackage{longtable}
\usepackage{multicol}

\usepackage{xcolor}
\usepackage[table]{xcolor}
\usepackage{soul}
\definecolor{darkgreen}{RGB}{84, 130, 53}
\definecolor{darkblue}{RGB}{46, 117, 182}
\definecolor{darkred}{RGB}{192, 0, 0}
\definecolor{myhighlightcolor_gray}{RGB}{229,229,229}
\definecolor{myhighlightcolor_brown}{RGB}{245,237,230}

\usepackage{bm}
\usepackage{tcolorbox}
\usepackage{makecell}

\usepackage{tabularray}

\def \eg {\emph{e.g.}}
\def \ie {\emph{i.e.}}
\def \etal {\emph{et al. }}

\begin{document}

\title{MotionMERGE: A Multi-granular Framework for Human Motion Editing, Reasoning, Generation, and Explanation}

\author{
Bizhu Wu, 
Jinheng Xie, 
Wenting Chen, 
Zhe Kong,
Jianfeng Ren,~\IEEEmembership{Senior Member,~IEEE,}
Linlin Shen,~\IEEEmembership{Senior Member,~IEEE,}
Ruibin Bai,~\IEEEmembership{Senior Member,~IEEE,}
Rong Qu~\IEEEmembership{Fellow,~IEEE}
        

\thanks{
This work was supported in part by the National Key R\&D Program of China under Grant 2024YFF0618400, 
National Natural Science Foundation of China under Grant 62576216,  
Guangdong Provincial Key Laboratory under Grant 2023B1212060076, 
and by Ningbo Science \& Technology Bureau under Grant 2024Z110, 2024Z124 and 2025Z197.}
\thanks{B. Wu, and L. Shen are with the Computer Vision Institute, School of Computer Science and Software Engineering, Shenzhen University, Shenzhen 518060, China, 
also with the Guangdong Key Laboratory of Intelligent Information Processing, Shenzhen University, Shenzhen 518060, China, 
and also with the School of Computer Science, University of Nottingham Ningbo China. (email: wubizhu@email.szu.edu.cn; llshen@szu.edu.cn)}
\thanks{J. Xie is with the Department of Electrical and Computer Engineering, National University of Singapore (email: sierkinhane@gmail.com)}
\thanks{W. Chen is with the Department of Radiation Oncology, Stanford University (email: wentchen@stanford.edu)}
\thanks{Z. Kong is with the Sun Yat-sen University. (email: kongzhecn@gmail.com)}
\thanks{J. Ren and R. Bai are with the School of Computer Science, University of Nottingham Ningbo China (email: Jianfeng.Ren@nottingham.edu.cn; Ruibin.BAI@nottingham.edu.cn)}
\thanks{R. Qu is with the School of Computer Science, University of Nottingham, Nottingham, U.K. (e-mail:
rong.qu@nottingham.ac.uk)}
\thanks{\dag: J. Ren and L. Shen are the corresponding authors.}
}

\markboth{IEEE Transactions on Pattern Analysis and Machine Intelligence}%
{Shell \MakeLowercase{IEEE Transactions on Pattern Analysis and Machine Intelligence}}



\maketitle

\begin{abstract}
Recent motion-language models unify tasks like comprehension and generation but operate at a coarse granularity, lacking fine-grained understanding and nuanced control over body parts needed for animation or interaction.
This stems from fundamental issues in both the model and the data, in which the model can't focus on motion’s localized pattern, and the training data lacks fine-grained supervision.
To tackle this, we propose MotionMERGE, a unified framework that bridges the granularity gap. 
First, we pioneer the study of fine-grained language-guided motion control, including detailed understanding and localized editing, by explicitly modeling motion at part and temporal levels within a single LLM, thereby endowing the model with robust priors for precise control. 
Second, we design Reasoning-Aware Granularity-Synergy pre-training, a novel strategy that employs joint supervision for cross-granularity alignment, temporal grounding, localized alignment, motion coherency, and motion-grounded chain-of-thought (CoT) reasoning. This equips the model with fine-grained motion–language alignment, cross-granularity synergy, and explicit reasoning ability. 
Third, we curate MotionFineEdit, a large-scale dataset (837K atomic + 144K complex triplets) with the first fine-grained spatio-temporal corrective instructions and motion-grounded CoT annotations, establishing a new benchmark for fine-grained text-driven motion editing and motion-grounded reasoning.
Extensive experiments demonstrate the capability of MotionMERGE for more precise motion generation, understanding, and editing, and compelling zero-shot generalization to other complex motion tasks. 
This work represents a significant step toward models that interact with motion in finer granularity and human-like reasoning.
\end{abstract}




\begin{figure}[!t]
    \centering
    \includegraphics[width=0.95\linewidth]{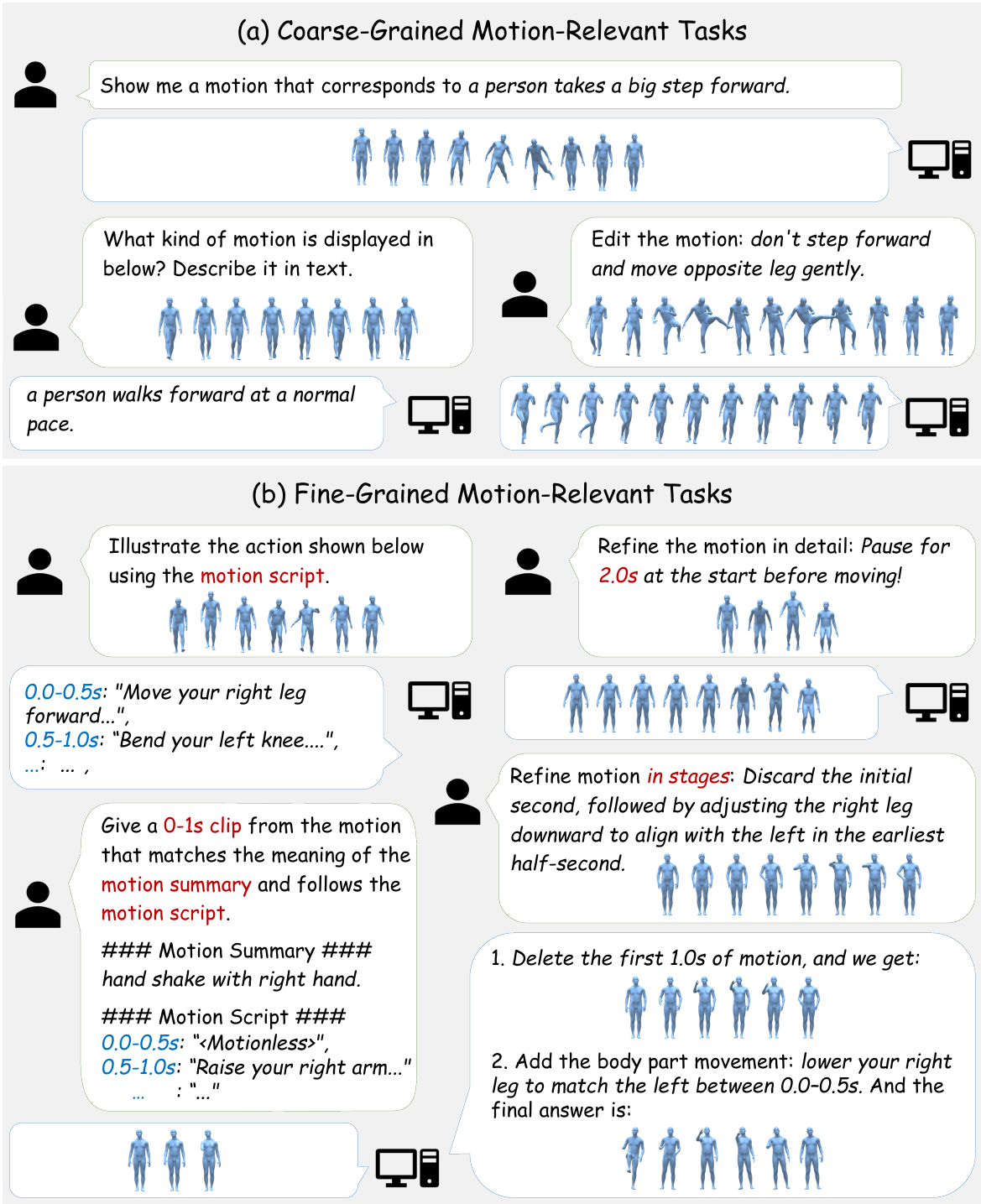}
    \caption{
        MotionMERGE unifies diverse motion-related tasks across granularities. 
        The \textit{upper} block showcases its results of coarse-grained tasks, including motion captioning (understanding), text-to-motion generation, and coarse motion editing. 
        The \textit{bottom} block highlights its fine-grained motion modeling, including detailed motion captioning, motion localization, and atomic/complex fine-grained motion editing. 
        Motions progress left-to-right (snapshots every 0.5s). 
        \textcolor{darkgreen}{Green}: inputs; 
        \textcolor{darkblue}{Blue}: outputs.
    }
    \label{fig:intro}
\end{figure}

\section{Introduction}
\label{sec:intro}

\IEEEPARstart{H}{uman} motion modeling is fundamental to applications in AR/VR, gaming, and virtual environments~\cite{wu2024motionllm, mogen_survey}. 
The adoption of natural language as an intuitive interface has spurred progress in motion-related tasks, enabling more interactive systems~\cite{openmotionlab_motiongpt}.
Existing motion-language research progresses along two axes: task specificity and model unification. 
Early specialized models target individual tasks, including motion captioning~\cite{guo2022tm2t}, text-to-motion generation~\cite{mdm, t2mgpt, humanml3d, guo2024momask}, and text-driven motion editing~\cite{athanasiou2024motionfix, kim2023flame, mdm, tevet2022motionclip, tseng2023edge}. 
These isolated designs, however, employ distinct architectures and training schemes, hindering cross-task generalization.

Recently, unified frameworks such as MotionGPT~\cite{openmotionlab_motiongpt}, MotionLLM~\cite{wu2024motionllm}, and MotionLab~\cite{guo2025motionlab} have emerged to alleviate this limitation. By leveraging shared representations, they handle multiple tasks simultaneously, improving cross-task transferability and reducing deployment costs. 
Yet, as summarized in Tab.~\ref{table:motion_llm_comparison}, these unified models remain coarse‑grained, treating motion as a monolithic sequence and using concise textual descriptions to record its overall semantics (see \textit{upper} block in Fig.~\ref{fig:intro}), thus lacking fine-grained control and the ability to understand, edit, and reason about specific body parts or temporal segments. 
This gap prevents them from being deployed in scenarios requiring high-precision, spatio-temporally localized control, such as animation, rehabilitation, and human–computer interaction.

We attribute this limitation to a dual bottleneck: insufficient modeling of localized motion patterns and a lack of fine-grained supervision. Addressing this requires both rich annotated data and a model capable of joint global-local semantic modeling. 
To tackle this, we introduce 
\textbf{MotionMERGE} (\textbf{M}ulti-granular motion \textbf{E}diting, \textbf{R}easoning, \textbf{G}eneration, and \textbf{E}xplanation, \ie, comprehension), 
a holistic framework that bridges the granularity gap via three synergistic innovations: 
a unified multi-granular motion-language model,
a reasoning-aware granularity-synergy pretraining strategy,
and a large-scale fine-grained motion editing dataset.

Specifically, we instantiate this framework by integrating a Motion VQ-VAE to discretize motion into movement primitive tokens, and a Motion-Aware Language Model that constructs a unified vocabulary merging text, motion, and special tokens (\eg, for temporal separation and motionless states). This design allows us to reformulate diverse motion tasks as conditional text generation within a single LLM. Crucially, this model bridges the granularity gap by enabling dense, part-level supervision and joint global-local alignment via next-token prediction, reducing the instability of joint low-level infilling and high-level reasoning.

\begin{table}[!t]
\caption{
    Comparison of recent motion–language unified frameworks. Supported tasks: \textbf{Gen.} (text-to-motion), \textbf{Und.} (motion-to-text), \textbf{Edit.} (motion editing), \textbf{Reas.} (motion reasoning). Granularity: \textbf{C} (coarse), \textbf{F} (fine).
} 
\label{table:motion_llm_comparison}
\begin{center}
\setlength{\tabcolsep}{8pt} 
\begin{tabular}{lcccc}
    \toprule[1pt]
    Method & Gen. & Und. & Edit. & Reas. \\ 
    \midrule[0.5pt]
    MotionGPT~\cite{openmotionlab_motiongpt} $_\text{NeurIPS’23}$ & C & C & - & - \\
    M$^3$GPT~\cite{luo2024m3gpt} $_\text{NeurIPS’24}$ & C & C &  \\
    MotionLLM~\cite{wu2024motionllm} $_\text{ArXiv’24}$ & C & C & - & - \\
    MotionGPT~\cite{zhang2024motiongpt} $_\text{AAAI’24}$ & C & - & - & - \\
    MG-MotionLLM~\cite{mgmotionllm} $_\text{CVPR'25}$ & C+F & C+F & - & -\\
    MotionLab~\cite{guo2025motionlab} $_\text{ICCV'25}$ & C & C & C & - \\
    \midrule[0.5pt]
    \textbf{MotionMERGE (Ours)} & C+F & C+F & C+F & C+F \\
    \bottomrule[1pt]  
\end{tabular}
\end{center}
\end{table}

However, naïvely training on mixed-granularity data underperforms due to fundamental misalignment between discrete motion tokens and multi-level semantics, compounded by limited reasoning capacity from direct input–output supervision. We resolve this with \textbf{R}easoning-\textbf{A}ware \textbf{G}ranularity-\textbf{S}ynergy (\textbf{RAGS}), a pre-training strategy that establishes a structured paradigm beyond conventional isolated task supervision. Instead of treating different objectives separately, RAGS strategically integrates four targeted supervisions: 
1)~\textit{temporal-aware supervision} that anchors motion tokens or fine-grained descriptions to precise intervals, overcoming the LLM’s lack of temporal priors; 
2)~\textit{localized alignment supervision} that explicitly learns partial fine-grained motion–language correspondences, easing fine-grained alignment and closing the gap between coarse- and fine-grained descriptions;
3)~\textit{motion coherent supervision} that introduces sparse motion tokens to reframe generation as infilling, enhancing the coherent dependencies across motion tokens; and 
4)~\textit{motion-grounded chain-of-thought supervision} that decomposes complex motion transformations into sequences of intermediate states, transcending direct input–output mapping. 
Together, these designs elevate alignment into structured reasoning and cross-granularity synergy, which effectively address the \textit{modeling} bottleneck of existing unified motion-language frameworks.

To enable fine-grained motion editing and reasoning, we introduce \textbf{MotionFineEdit}, a large-scale dataset comprising approximately \textbf{837K} atomic editing triplets and \textbf{144K} complex triplets annotated with motion-grounded Chain-of-Thought (CoT), which helps addressing the key \textit{data} bottleneck for unified frameworks.
Each triplet provides a source motion, a target motion, and fine-grained correctional textual instructions (Fig.~\ref{fig:data_example}).
MotionFineEdit advances prior datasets~\cite{athanasiou2024motionfix, jiang2025motionrefit} in two critical dimensions. 
First, it supports \textit{precise spatio-temporal control}, with annotations for body-part level editing, explicit temporal intervals, and duration modification, moving beyond coarse, whole-body edits. 
Second, it provides the first \textit{motion-grounded CoT annotations}, decomposing complex edits into sequences of intermediate motion states to facilitate interpretable, stepwise reasoning beyond traditional text-based CoT. 
To ensure high-quality annotations, we developed a scalable construction pipeline that integrates automatic generation with rigorous human verification, enforcing consistency and fidelity across all samples. 
Beyond supporting our training, MotionFineEdit establishes a challenging benchmark that starkly reveals the \textit{granularity gap}: state-of-the-art methods exhibit a drastic performance drop on it (\eg, R@1 of~\cite{athanasiou2024motionfix} falls from 35.16\%to 9.32\%), underscoring the necessity of the newly proposed fine-grained motion editing.

This work extends our previous MG-MotionLLM~\cite{mgmotionllm} (CVPR'25) from a model unifying generation and understanding to a fully unified framework that additionally supports multi-granular motion editing and reasoning.
MotionMERGE achieves this by four key advances:  
1)~an enhanced unified framework supporting generation, understanding, editing, and reasoning;
2)~a new RAGS (Reasoning-Aware Granularity-Synergy) pretraining strategy; 
3)~a new fine-grained text-driven human motion editing task, together with a scalable data construction pipeline that yields the large-scale MotionFineEdit dataset; and 
4)~an innovative technique for constructing the first motion-grounded CoT supervision. 
The result is a more general motion–language modeling paradigm, validated by superior performance across tasks.

Our key contributions are fourfold: 
1)~We pioneer the study of text-driven fine-grained motion-language tasks (\eg, comprehension, editing, and reasoning), and propose MotionMERGE, the first framework to unify diverse motion-language tasks across multiple granularities within a single model. 
2)~We introduce RAGS pre-training, a novel strategy that employs joint supervision for cross-granularity, temporal grounding, localized alignment, motion consistency, and motion-grounded chain-of-thought reasoning, enhancing effective learning across multiple granularities with improved controllability and interpretability. 
3)~We construct MotionFineEdit, a dataset featuring fine-grained spatio-temporal corrective instructions and motion-grounded CoT annotations. Notably, it serves as a challenging benchmark that reveals the granularity gap in existing methods, and is built via a scalable semi-automatic pipeline designed to minimize annotation effort. 
4)~Extensive experiments validate the effectiveness of MotionMERGE. It outperforms state-of-the-art methods on both coarse- and fine-grained benchmarks, supports complex editing scenarios, and demonstrates zero-shot generalization of reasoning capabilities across tasks.

\section{Related Work}
\label{sec:related_work}

\subsection{Text-Driven Human Motion Generation}
\label{sec:related_work_t2m}
Human motion generation can be conditioned on diverse inputs: action labels~\cite{actor, raab2023modi, tevet2022motionclip}, music~\cite{li2022danceformer, li2025lodge++, yang2025lagrangian, MNET}, speech~\cite{xu2025combo, li2023audio2gestures, ng2024audio} and text~\cite{t2mgpt, mdm, petrovich2022temos, DrawMotion}. 
Among these, natural language has emerged as a particularly powerful and flexible modality due to its expressiveness and intuitive interface, driving significant progress in text-to-motion generation~\cite{kim2023flame, barquero2024flowmdm, huang2024como, petrovich24stmc, wang2023fgt2m, lamp, lu2025scamo, Chen_2025_CVPR, parco}.  
The prevailing methodologies can be categorized into three technical paradigms, each with distinct trade-offs. 
\textit{Diffusion-based} methods~\cite{zhang2024motiondiffuse, mdm, chen2023mld} learn a reverse denoising process to generate high-fidelity motion sequences, often achieving superior sample quality at the expense of slower, iterative sampling. 
\textit{Autoregressive} approaches~\cite{t2mgpt, guo2024momask, parco} frame generation as sequential prediction of discretized motion tokens, enabling faster inference but potentially suffering from error accumulation in long sequences.
To balance these trade-offs, \textit{hybrid} methods~\cite{AMD, CLoSD, MotionStreamer} have emerged, which efficiently generate segments in an autoregressive manner while refining them locally using diffusion-based techniques.

Despite these advances, a fundamental limitation persists across paradigms: reliance on \textit{coarse-grained textual descriptions} (\eg, from HumanML3D~\cite{humanml3d} and KIT-ML~\cite{kitml}) that summarize global actions but omit the \textit{fine-grained kinetic details} of body-part movements, hindering precise spatio-temporal alignment. 
Recent efforts to incorporate finer annotations include using LLM-based enrichment~\cite{kalakonda2023actiongpt, fgt2m++}, crafting more accurate descriptions~\cite{he2023semanticboost, yazdian2023motionscript}, and curating dedicated fine-grained datasets~\cite{zhang2023finemogen, finemotion}. 
While these improve generation precision, they remain largely confined to the generation task, overlooking the synergistic potential of jointly modeling fine-grained motion understanding, generation, and reasoning.

\subsection{Human Motion Editing}
Human motion editing aims to modify motion sequences to satisfy user requirements, which is essential for animation and interactive applications. 
Early research primarily focused on \textit{whole-body editing}, such as transferring style or amplitude, where entire motions are transformed to match target styles. 
These methods often relied on supervised learning with paired data~\cite{xia2015realtime_styleTrans}, while more recent approaches relaxed this constraint through unpaired training paradigms~\cite{aberman2020unpaired_styleTrans, jang2022motion_styleTrans, tao2022_styleTrans}. 
Methods like MotionCLIP~\cite{tevet2022motionclip} and SALAD~\cite{hong2025salad} further enabled global adjustments via latent space or attention-based manipulations. 
However, these approaches operate at global level and lack fine-grained spatial control.

To achieve localized control, recent methods have shifted towards \textit{part-based editing}. 
A dominant strategy formulates editing as a motion inpainting problem within diffusion models, where text conditions the regeneration of specific body regions while preserving the rest~\cite{mdm, zhang2024motiondiffuse, kim2023flame, pinyoanuntapong2024controlmm}. 
Other works, such as CoMo~\cite{huang2024como} and FineMoGen~\cite{zhang2023finemogen}, incorporate LLMs to interpret high-level user instructions into part-specific descriptions. 
Iterative Motion Editing~\cite{goel2024iterative} decomposes the process by first using an LLM to map instructions to relevant joints and frames, followed by diffusion-based infilling. 
While improving spatial controllability, these methods frequently depend on predefined joint masks or explicit specifications (\eg, which joints and when to edit), which places a burden on the user and limits applicability in scenarios where editing intent is best expressed in natural language.

This gap has motivated the exploration of text-driven motion editing as a more flexible and intuitive interface. 
Pioneering work like MotionFix~\cite{athanasiou2024motionfix} introduced a dataset of source-target motion pairs annotated with textual descriptions of the change, alongside a diffusion-based editing model. 
Subsequent efforts, including SimMotionEdit~\cite{li2025simmotionedit} and MotionLab~\cite{guo2025motionlab}, have improved editing fidelity through auxiliary training objectives and multi-task learning. 
Dynamic Motion Blending~\cite{jiang2025motionrefit} utilizes a hybrid autoregressive-diffusion framework supported by novel data augmentation, enabling a broader range of edits. 

Despite recent progress, text-driven motion editing remains largely coarse, supporting whole-body modifications with limited temporal control. 
This limitation originates from a dual bottleneck: model that lack explicit mechanisms for fine-grained spatio-temporal grounding, and the absence of large-scale datasets providing part-level and temporally-localized annotations. 
To overcome this, we introduce MotionFineEdit, a dataset for fine-grained editing with body-part and temporal control and motion-grounded CoT; and MotionMERGE, a unified framework that leverages this data for spatio-temporally precise motion editing.

\subsection{Human Motion Reasoning} 
While LLMs excel at comprehension and generation, their higher-level reasoning remains limited~\cite{li2023weakly, xiong2024teilp, Reasoning_survey}, spurring significant research efforts~\cite{wei2022cot, ding2025eagle, AtomThink, EvolveNav}. 
In the motion domain, however, research on reasoning remains sparse and highly constrained. 
Existing works can be categorized into two primary paradigms. 
1)~Early methods such as NSPose~\cite{endo2023NSPose} and IMoRe~\cite{li2025IMoRe} \textit{explicitly} model reasoning by decomposing motion understanding into symbolic programs or structured memory. While enabling multi-step inference, their reliance on hand-crafted modules prevents end-to-end learning in a unified framework, limiting flexibility and scalability. 
Recent methods circumvent the lack of motion-grounded data by leveraging language-only LLM. They use LLMs (\eg, ChatGPT~\cite{openai_chatgpt} and GPT-4~\cite{gpt4}) to infer related information~\cite{jiang2024motionchain, zhou2024avatargpt, park2025MoLaM} (\eg, contextual scenes) or decompose instructions into multiple calls to a motion-text translation model based solely on captions~\cite{motionagent}. 
This may leads to a reasoning–motion dissociation: LLM-derived steps (\eg, ``stand, then walk, then stand'') often misalign with actual motions (\eg, ``walk''). 
2)~Another line employs feedback-driven optimization (\eg, human feedback, reinforcement learning) to \textit{implicitly} encourage reasoning for better alignment~\cite{han2024hutumotion, wang2024MotionCritic, mao2024InstructMotion, liu2024motionrl, ouyang2025motionr1}. 
While improving generation, these methods suffer from unstable training and poor interpretability.

In contrast, we pioneer motion-grounded Chain-of-Thought (CoT) data for fine-grained editing. Each CoT trace provides intermediate motion states aligned with decomposed steps, offering high-fidelity, interpretable supervision. This enables our unified framework to perform accurate, explicit reasoning, improving generalization across motion tasks.

\subsection{Motion Large Language Models}
The success of large language models in linguistic tasks~\cite{dubey2024llama3, t5, gpt4} has spurred their extension to multimodal understanding and generation, giving rise to unified models for vision~\cite{liu2023visual, sd3, wu2024next, xie2024show, zhou2024transfusion}, video~\cite{zhang2023videollama, xie2025showo2}, and audio~\cite{deshmukh2023pengi, hurst2024gpt4o, whisper}. This progress has naturally inspired the integration of human motion as a ``language'' within LLMs~\cite{chen2025motionllm, openmotionlab_motiongpt, luo2024m3gpt}, leading to a new paradigm: \textit{Motion LLMs}. These models represent motion as discrete tokens and process them alongside text via a shared LLM backbone, aiming to unify diverse motion–language tasks. 
Pioneering works like MotionGPTs~\cite{openmotionlab_motiongpt, zhang2024motiongpt} demonstrated that motion tokens could be incorporated into models like LLaMA and T5, enabling a single model to perform generation, captioning, and prediction. Subsequent efforts expanded this unified framework to multi-human scenarios~\cite{wu2024motionllm} and incorporated additional modalities like music for richer cross-modal mapping~\cite{luo2024m3gpt}. These works collectively establish a promising unified modeling paradigm, moving beyond task-specific architectures.

However, this unified paradigm encounters a representational bottleneck: it focuses on the sequence-level alignment, where one concise sentence describes an entire motion~\cite{openmotionlab_motiongpt, wu2024motionllm, luo2024m3gpt}. This coarse alignment fails to capture motion’s detailed patterns, limiting spatial-temporal grounding and controllability. Advancing the paradigm therefore requires moving beyond sequence-level supervision. In this work, we introduce structured, fine-grained supervision that explicitly aligns language with body parts and temporal intervals, equipping the LLM framework with the necessary inductive biases for precise understanding and control, elevating Motion LLMs beyond the current coarse-grained regime.

\section{Proposed MotionMERGE}
\begin{figure*}[!t]
\begin{center}
\includegraphics[width=1.0\linewidth]{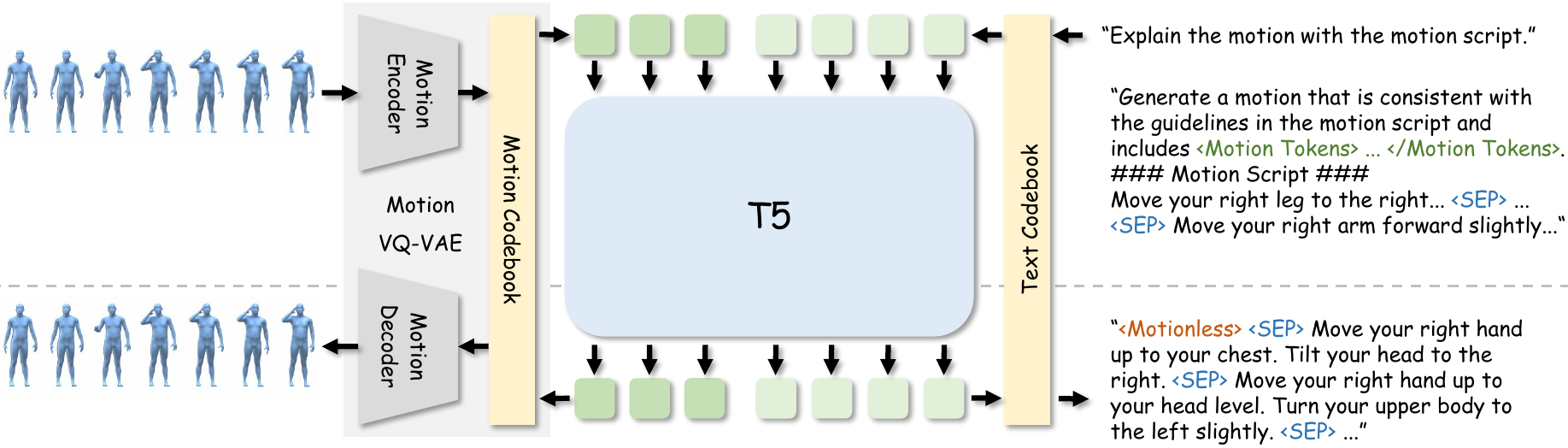}
\end{center}
\vspace{-10pt}
\caption{
    \textbf{Overview of MotionMERGE.}
    Motions are converted into special text, allowing all tasks to be formulated as conditional text generation. The framework comprises a motion VQ-VAE that transforms continuous motion into discrete tokens, and a T5-based language model that processes interleaved text and motion tokens. It explicitly handles diverse motion-language tasks (\eg, generation, editing) at both global and local (body-part, temporal) levels within a unified framework through different promptings, enabled by joint training on multi-granularity data.
}
\label{fig:fine_motion_llm}
\end{figure*}

\subsection{Overview of MotionMERGE}
We propose \textbf{MotionMERGE}, a unified motion-language framework that addresses diverse tasks (\textit{generation, editing, comprehension, and reasoning}) across multiple granularities within a single model. 
As illustrated in Fig.~\ref{fig:fine_motion_llm}, MotionMERGE achieves granularity-agnostic unification by encoding motion into discrete tokens, indicating granularity via text, and reformulating all tasks as conditional text generation within a single LLM.
This token-based formulation reduces computational costs, eases error accumulation during inference, and effectively unifies disparate granularities. 
Specifically, it comprises two core components: 1) a Motion VQ-VAE (Sec.~\ref{sec:motion_vqvae}) that encodes continuous motion into a discrete token sequence and decodes it back; and 2) a motion-aware language model (Sec.~\ref{sec:motion_llm}), built upon a T5 backbone, that processes interleaved motion and text tokens.
The model is trained via a two-stage scheme: Reasoning-Aware Granularity-Synergy (RAGS) pre-training (Sec.~\ref{sec:rags}), which enables it to master this unified, instruction-driven paradigm, followed by task-specific instruction tuning for optimal adaptation.

\subsection{Motion VQ-VAE}
\label{sec:motion_vqvae}
Raw motion data is inherently high-dimensional, and directly prompting a language model to generate such dense numerical sequences incurs substantial computational cost while increasing the risk of error accumulation during inference. 
To address this issue, we employ a Vector Quantized Variational Autoencoder (VQ-VAE)~\cite{van2017vqvae} to learn discrete latent representations. Specifically, we adopt the motion VQ-VAE from T2M-GPT~\cite{t2mgpt}, which has demonstrated effectiveness in capturing human motion dynamics for text-conditioned generation. This provides a robust, off-the-shelf tokenizer that maps motion to a finite vocabulary, forming the essential interface between the continuous motion space and the discrete processing of our language model.

The motion VQ-VAE comprises an encoder $\mathcal{E}$, a decoder $\mathcal{D}$, and a learnable codebook $\bm{B}=\{\bm{b}_i\}_{i=1}^{N_B}$, where $N_B$ is the codebook size. 
Given a $T$-frame motion sequence $\bm{M} = [\bm{m}_1, \bm{m}_2, \dots, \bm{m}_T]$, where $\bm{m}_t \in \mathbb{R}^{d_m}$, 
the encoder $\mathcal{E}$ maps it into latent features 
\[
\bm{Z} = \mathcal{E}(\bm{M}) = [\bm{z}_1, \bm{z}_2, \dots, \bm{z}_{T/l}], \quad \bm{z}_t \in \mathbb{R}^{d_z},
\]
where $l$ is the temporal downsampling factor. 
Each latent feature is then quantized into a motion token
\begin{equation}
    c_t=\underset{i \in \{ 1, \dots, N_B\}}{\arg\min }\left\|\bm{z}_t-\bm{b}_i\right\|_2^2 ,
\end{equation}
resulting in a motion token sequence $\bm{c} = [c_1, c_2, \dots, c_{T/l}]$.

Conversely, given a sequence of motion tokens $\bm{c}$, we first retrieve the corresponding quantized latent features from the codebook $\hat{\bm{Z}} = [\hat{\bm{z}}_1, \hat{\bm{z}}_2, \dots, \hat{\bm{z}}_{T/l}]$ with $\hat{\bm{z}}_i = \bm{b}_{c_i}$.
Then, the decoder $\mathcal{D}$ reconstructs $\hat{\bm{Z}}$ into a motion sequence 
\begin{equation}
    \hat{\bm{M}} = \mathcal{D}(\hat{\bm{Z}}) = [\hat{\bm{m}}_1, \hat{\bm{m}}_2, \dots, \hat{\bm{m}}_T].
\end{equation}
 
The motion VQ-VAE is optimized with reconstruction, embedding, and commitment losses:
\begin{equation}
    \begin{aligned}
        \mathcal{L}_{\text{VQ-VAE}} =
        \ \|\bm{M}-\hat{\bm{M}}\|_2^2 
        &+ \|\text{SG}(\bm{Z})-\hat{\bm{Z}}\|_2^2 \\
        &+ \beta \|\bm{Z}-\text{SG}(\hat{\bm{Z}})\|_2^2 ,
    \end{aligned}
\end{equation}
where $\beta$ controls the commitment strength and $\text{SG}(\cdot)$ denotes stop-gradient.

After training, the motion VQ-VAE is frozen and provides a stable discrete vocabulary, allowing the language model to focus on cross-modal alignment and reasoning while avoiding the instability of jointly optimizing low-level reconstruction.

\subsection{Motion-Aware Language Model}
\label{sec:motion_llm}
The core of our unified framework is a motion-aware language model that processes interleaved sequences of text and motion tokens. To enable this, we first construct a unified vocabulary $\bm{V}$ that merges three disjoint sets: 
1)~The original text vocabulary $\bm{V}_\text{t}=\{v_\text{t}^i\}_{i=1}^{N_\text{t}}$; 
2)~A motion vocabulary $\bm{V}_\text{m}=\{\textless 1\textgreater, \textless 2\textgreater, \dots, \textless N_B\textgreater\}$ corresponding to the VQ-VAE codebook indices; 
3)~A set of \textit{special tokens} $\bm{V}_\text{s}$ that provides structural and control signals for the multimodal sequences, enabling fine-grained control:
\begin{itemize}
    \item \textcolor{magenta}{\textless Motion Tokens\textgreater} and \textcolor{magenta}{\textless /Motion Tokens\textgreater} delimit motion token sequences, allowing the model to clearly distinguish motion from text.
    \item \textcolor{cyan}{\textless SEP\textgreater} separates descriptions of different motion snippets, which is essential for representing fine-grained body part movements.
    \item \textcolor{orange}{\textless Motionless\textgreater} explicitly denotes motion snippets with no significant body part movement, providing a dedicated marker to facilitate precise temporal alignment between fine-grained descriptions and motion.
\end{itemize}
This unified vocabulary $\bm{V} = \bm{V}_\text{t} \cup \bm{V}_\text{m} \cup \bm{V}_\text{s}$ allows all motion-related tasks to be expressed within a single token space, where task and granularity are dictated by the sequence structure and special tokens, rather than separate model heads.

Following the success of encoder-decoder transformers in multimodal unification~\cite{t5, openmotionlab_motiongpt}, we adopt T5~\cite{t5} as our backbone LLM. 
Given an input token sequence $\bm{X}_\text{in} = \{v_\text{in}^i \}_{i=1}^{N_\text{in}}$, the model generates an output sequence $\bm{X}_\text{out} = \{v_\text{out}^i \}_{i=1}^{N_\text{out}}$, where $v_\text{in}^i$ and $v_\text{out}^i$ are from $\bm{V}$, and $N_\text{in}$ and $N_\text{out}$ denote the numbers of input and output tokens. 
The model is trained via next-token prediction by minimizing the standard cross-entropy loss:
\begin{equation}
\label{eqn:llm}
    \mathcal{L}_{\text{LLM}} = -\sum_{i=1}^{N_\text{out}} \log(P(v_\text{out}^i \mid \bm{X}_\text{in}, v_\text{out}^{<i}, \theta_\text{LLM})).
\end{equation}
This enables the model to learn text-motion correspondences directly from data and, at inference, to flexibly execute multiple motion-related tasks within a single model based on different prompts.

\subsection{Reasoning-Aware Granularity-Synergy Pretraining}
\label{sec:rags}

To transform a plain LLM into a unified model for multi-granularity motion–language alignment,
we first establish basic cross-modal alignment by jointly fine-tuning the LLM on mixed-granularity data, which exposes two core deficiencies: misalignment between discrete motion tokens and multi-level semantics, and limited reasoning capacity from direct input–output supervision. 
To resolve these issues, we introduce Reasoning-Aware Granularity-Synergy (RAGS) Pre-training, a curriculum-inspired structured scheme that incorporates multiple targeted inductive biases. Optimized jointly beyond isolated multi-task learning, these biases create a synergistic effect that enables coherent cross-granularity learning.

\noindent\textbf{1)~Temporal-Aware Supervision:} \quad
Standard language models lack prior knowledge of time underlying motion tokens and detailed descriptions, a prerequisite for fine-grained understanding and control. We inject this bias via auxiliary tasks that explicitly require the model to align motion tokens and descriptions with their corresponding temporal intervals (\eg, extracting motion segments from specified intervals). This teaches the model temporal structure and detailed descriptions, enabling temporally precise operations. 

\noindent\textbf{2)~Localized Alignment Supervision:} \quad
Directly aligning lengthy fine-grained descriptions (often $>$1000 tokens) with short motion sequences ($<$50 tokens) creates an extreme length mismatch that obscures local correspondences. We bridge this divide by decomposing the problem: introducing tasks that align \textit{part of detailed descriptions} with their corresponding \textit{local motion segments}. This forces the model to establish fine-grained, part-level correspondences in manageable chunks, building a ``dictionary'' of local text-motion mappings that can be composed for full-sequence understanding.

\noindent\textbf{3)~Motion Coherent Supervision:} \quad 
Simple text–motion alignment provides weak structural constraints between motion tokens, limiting fine-grained motion coherence. We address this by introducing auxiliary tasks that additionally condition on sparse motion tokens, reframing generation as motion infilling. This reduces reliance on brittle direct text–motion alignment and encourages coherent dependencies across motion tokens, enhancing sequence-level plausibility.

\noindent\textbf{4)~Motion-Grounded CoT Supervision:} \quad
Prior motion LLMs suffer from either poor interpretability for complex tasks or language‑driven reasoning detached from actual motion. We address this by supervising the motion‑grounded CoT process, where each reasoning step is paired with an aligned intermediate motion state. Training the model to decompose complex instructions (\eg, ``\textit{first raise the left hand, then turn}'') into executable sub‑goals instills a composable, causal reasoning ability, improving both interpretability and motion alignment across diverse motion-language tasks.

Critically, these four supervisory signals are optimized \textit{jointly} within a single loss function, together with fine-tuning on motion–text pairs across granularities,
\begin{equation}
\label{eq:rags}
\mathcal{L}_{\text{RAGS}} = \mathcal{L}_{\text{gran}} + 
\lambda_\text{temp} \mathcal{L}_{\text{temp}} + 
\lambda_\text{local} \mathcal{L}_{\text{local}} + 
\lambda_\text{coh} \mathcal{L}_{\text{coh}} + 
\lambda_\text{cot} \mathcal{L}_{\text{cot}},
\end{equation}
where $\mathcal{L}_{\text{gran}}$ denotes the cross-entropy loss (Eq.~\ref{eqn:llm}) for fine-tuning on motion–text pairs across granularities, 
and $\mathcal{L}_{\text{temp}}$, $\mathcal{L}_{\text{local}}$, $\mathcal{L}_{\text{coh}}$, and $\mathcal{L}_{\text{cot}}$ are the cross-entropy losses for temporal grounding, localized alignment, motion coherent, and CoT tasks, respectively.
This joint optimization yields modeling that is both locally precise and globally aware. Overall, RAGS pre-training transforms the LLM from a model \textit{not even motion-aware} into one that \textit{well understands, manipulates, grounds, and reasons} about motion across granularities from language.

\section{MotionFineEdit Dataset}
\label{sec:dataset}
\begin{figure*}[!b]
    \centering
    \includegraphics[width=0.8\linewidth]{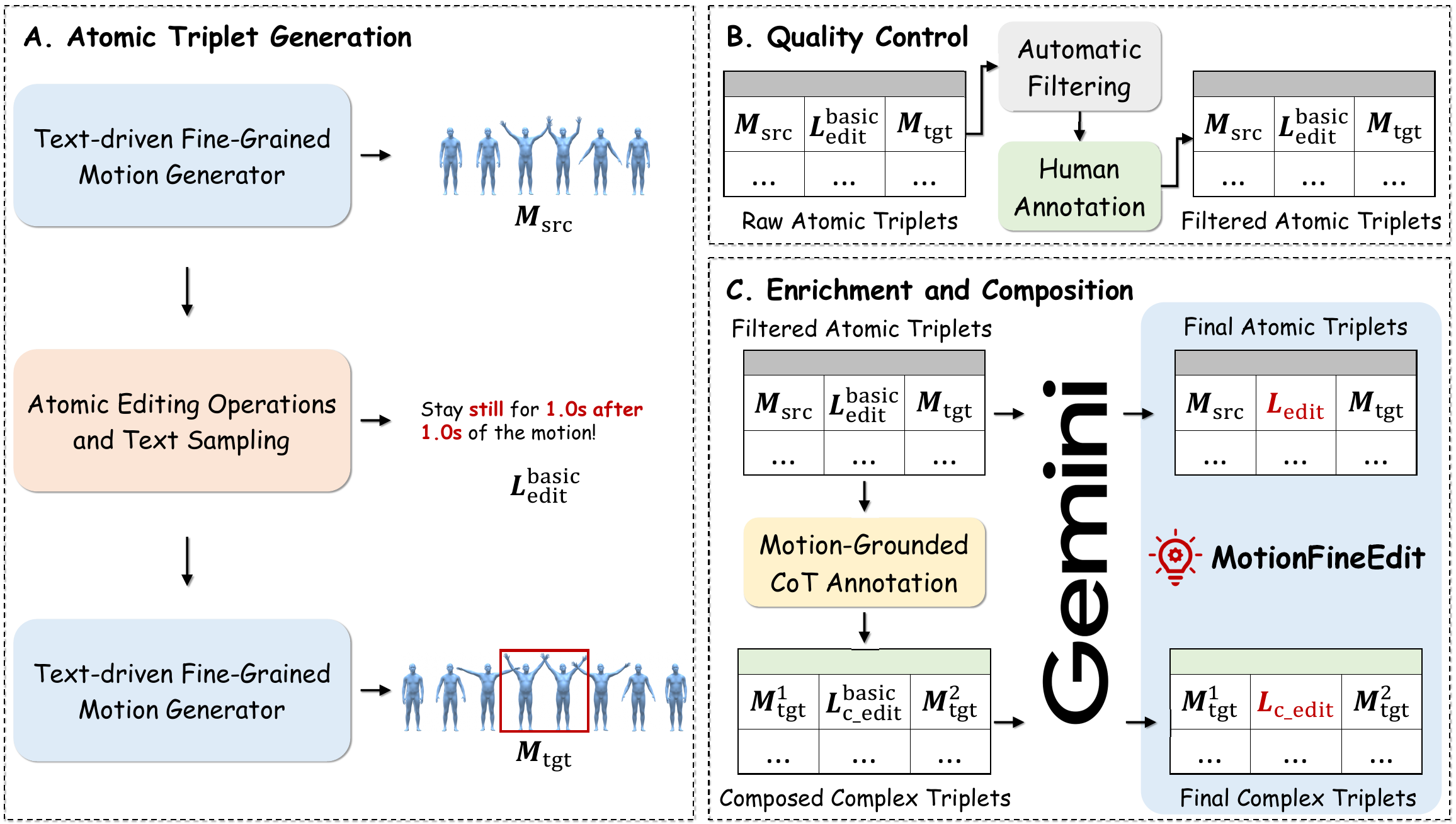}
    \caption{Construction pipeline of the MotionFineEdit dataset. 
    The pipeline consists of an atomic triplet stage, a quality control stage, and an enrichment and composition stage.
    Motions are snapshotted every 0.5s. 
    }
    \label{fig:data_pipeline}
\end{figure*}
Existing datasets for text-driven motion editing are limited to coarse, whole-body modifications and lack the fine-grained spatio-temporal annotations required for precise editing~\cite{athanasiou2024motionfix, jiang2025motionrefit}. To bridge this gap, we introduce MotionFineEdit, a large-scale dataset for text-driven fine-grained human motion editing. It provides part-level and temporally-localized textual annotations. Moreover, as an additional novel contribution, it introduces the first motion-grounded Chain-of-Thought annotations for interpretable, stepwise reasoning.

The dataset is constructed via a scalable three-stage pipeline (Fig.~\ref{fig:data_pipeline}):
1)~Atomic Triplet Generation (Sec.~\ref{sec:generation}) using a text-driven motion synthesizer; 
2)~rigorous Quality Control stage (Sec.~\ref{sec:qc_pipeline}) with automatic and human verification;
and 3) an Enrichment and Composition stage (Sec.~\ref{sec:enrichment}) for CoT construction and linguistic diversification. This structured approach ensures scalability and high fidelity of the dataset.

\subsection{Atomic Triplet Generation}
\label{sec:generation} 
Manually capturing motion pairs that differ only in specific body parts or precise timing is prohibitively difficult. 
We therefore adopt a generation-based paradigm, using a text-driven fine-grained motion generator $\mathcal{G}$ (Sec.~\ref{sec:fg_mogen}). 
This allows us to synthesize the source motion $\bm{M}_{\text{src}}$ and the target motion $\bm{M}_{\text{tgt}}$. 
To create an edit, we sample an atomic editing operation and produces a basic corrective instruction $L_{\text{edit}}^{\text{basic}}$ (Sec.~\ref{sec:fg_correctivetext_gen}), yielding a candidate triplet:
\begin{equation}
    (\bm{M}_{\text{src}}, L_{\text{edit}}^{\text{basic}}, \bm{M}_{\text{tgt}}). 
\end{equation}
This approach enables scalable production of fine-grained atomic editing triplets, which are subsequently validated by the quality control stage (Sec.~\ref{sec:qc_pipeline}).

\subsubsection{Text-driven Fine-Grained Motion Generator}
\label{sec:fg_mogen}
Producing fine-grained motion editing pairs $(\bm{M}_\text{src}, \bm{M}_\text{tgt})$ requires a generator that can be precisely controlled: the two motions must share identical global semantics, yet differ only in specific, localized spatio-temporal regions. Standard text-to-motion models, conditioned solely on coarse descriptions, cannot satisfy this requirement, as varying the text would alter the entire motion. 
To achieve this localized control, we employ a \textit{granularity-aware motion generator} $\mathcal{G}$ that accepts dual-conditioning inputs: a coarse action description $L_\text{c}$ (\eg, ``a person walks'') and a sequence of fine-grained, part-level temporal instructions $\bm{L}_\text{d} = [L_{\text{d}_1}, \dots, L_{\text{d}_k}]$, where each $L_{\text{d}_i}$ describes the movements of specific body parts over a 10-frame snippet. This design decouples global action context ($L_\text{c}$) from local kinematics ($\bm{L}_\text{d}$). We instantiate $\mathcal{G}$ by fine-tuning the pretrained model from MG-MotionLLM~\cite{mgmotionllm} (see Appendix for details), which is inherently architected for such hierarchical conditioning: $\bm{M} = \mathcal{G}(L_\text{c}, \bm{L}_\text{d})$.

To create a source–target pair, we first sample a coarse motion caption $L_\text{c}$ from HumanML3D~\cite{humanml3d} and retrieve its aligned fine-grained descriptions $\bm{L}_\text{d}^\text{src}$ from FineMotion~\cite{finemotion}. The source motion is $\bm{M}_\text{src} = \mathcal{G}(L_\text{c}, \bm{L}_\text{d}^\text{src})$. An atomic edit (Sec.~\ref{sec:fg_correctivetext_gen}) modifies $\bm{L}_\text{d}^\text{src}$ to $\bm{L}_\text{d}^\text{tgt}$, altering only the descriptions of the targeted body parts and/or time intervals. The target motion is then $\bm{M}_\text{tgt} = \mathcal{G}(L_\text{c}, \bm{L}_\text{d}^\text{tgt})$. Crucially, keeping $L_\text{c}$ unchanged ensures the global action remains consistent; the edit is induced solely through the difference between $\bm{L}_\text{d}^\text{tgt}$ and $\bm{L}_\text{d}^\text{src}$, enabling precise, fine-grained modifications.

Notably, such re-generation of the target motion may introduce unintended changes outside the specified edited regions. These candidate pairs are hence passed to the quality control pipeline (Sec.~\ref{sec:qc_pipeline}) for validation and filtering.

\subsubsection{Atomic Editing Operations and Text Sampling}
\label{sec:fg_correctivetext_gen}

\begin{figure*}[!b]
\centering
\includegraphics[width=1.0\linewidth]{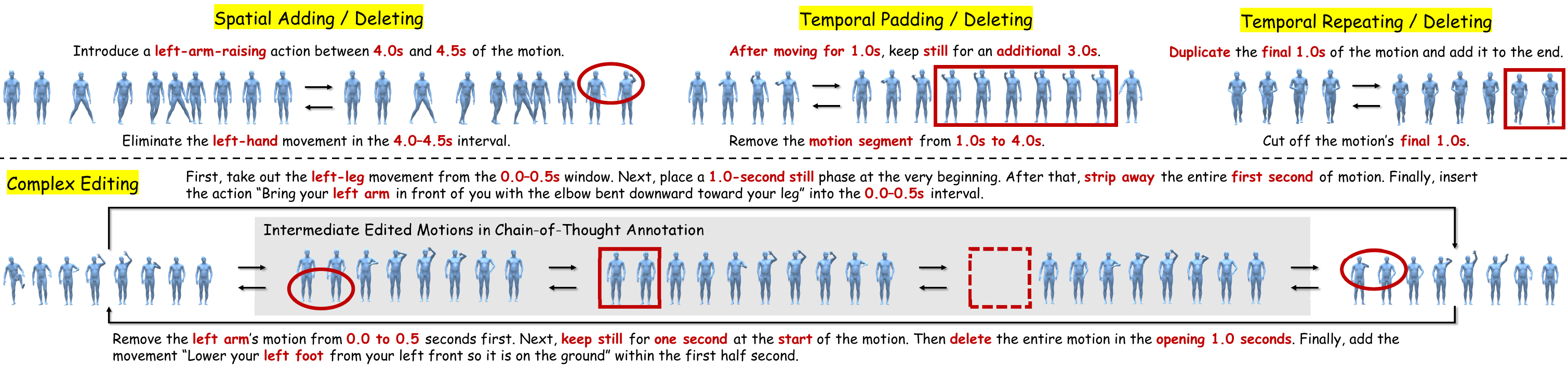}
\caption{
\textbf{Examples of text-driven fine-grained human motion editing from the MotionFineEdit dataset.}
Top: \textit{Atomic} edits targeting spatial or temporal dimensions.
Bottom: \textit{Complex} edit with chain-of-thought annotations, illustrating intermediate states across both dimensions.
Colored text marks temporal intervals and body parts; circles indicate edited body parts, rectangles edited time segments. Motions are sampled at 0.5s intervals.
}
\label{fig:data_example}
\end{figure*} 

To generate the target motion $\bm{M}_\text{tgt}$ and its corresponding corrective instruction $L_{\text{edit}}^{\text{basic}}$, we modify the source fine-grained description $\bm{L}_\text{d}^\text{src}$ in a \textit{controlled, interpretable, and composable} manner. 
The core challenge is to define a set of editing primitives that are expressive enough to cover common editing intents, produce unambiguous changes to $\bm{L}_\text{d}^\text{src}$ that can be reliably executed by the motion generator $\mathcal{G}$, and support the composition into complex, multi-step edits for reasoning (Sec.~\ref{sec:cot}). 

To address this, we introduce 11 atomic operations, factorized along two axes: \textit{scope} (where/when) and \textit{type} (what change). Temporal edits modify timing at the start, middle, or end via padding, repeating, or deleting. Spatial edits add or delete body-part movements within a given temporal snippet. This factorization ensures each operation is \textit{simple, unambiguous, and critically invertible}. The complete set provides a basis for the editing intents in this work. 

For each operation, we sample its parameters (\eg, the start index of motion snippets to edit $p$, their length $n$, the edited body-part movement description $s_\text{bpm}$) to produce a modified fine-grained description $\bm{L}_\text{d}^\text{tgt}$, and fit these parameters into templates to generate a basic corrective instruction $L_{\text{edit}}^{\text{basic}}$ that states the \textit{what}, \textit{where}, and \textit{when} of the edit. For instance, in \textit{temporal repeating}, a segment of $n$ snippets starting at $p$ is duplicated, and the instruction specifies the repeated interval. In \textit{spatial adding}, a new body-part movement $s_\text{bpm}$ is inserted into snippet $p$, with the instruction specifying the movement and its temporal interval. More details are given in Appendix. 

This systematic approach transforms an abstract edit into a precise modification to $\bm{L}_\text{d}^\text{src}$ and a human-readable instruction, enabling scalable generation of fine-grained editing triplets.

\subsection{Quality Control Pipeline}
\label{sec:qc_pipeline}
Candidate triplets from the generation stage may contain inaccuracies due to the stochastic nature of the motion generator. Ensuring data fidelity therefore requires a rigorous quality control (QC) process. We employ a \textit{human-in-the-loop pipeline} that synergizes the efficiency of automatic pre-filtering with the nuanced judgment of human verifiers. This design balances scalability with the high-quality standards required for fine-grained editing data. 

\subsubsection{Automatic Filtering}
\label{sec:auto_filter}
The first stage applies efficient, rule-based filters to prune obviously invalid candidates, reducing the workload for human annotators. We use a pretrained motion encoder $\mathcal{F}(\cdot)$~\cite{humanml3d} to compute motion similarities. Filters discard pairs that violate basic constraints, such as incorrect temporal length changes or unintended left-right mirroring. More importantly, for each atomic operation type, we define customized criteria that ensure the target motion $\bm{M}_\text{tgt}$ adheres to the intended edit relative to $\bm{M}_\text{src}$. For example, for spatial adding, we enforce that changes are confined to the specified body part; for temporal repeating, we ensure the correct segments are duplicated and unedited regions are preserved. This stage efficiently removes a large number of low-quality samples, passing a cleaner set to human verification. More details are given in Appendix.

\subsubsection{Human Annotation}
\label{sec:human_anno}
Automatic filters cannot capture all nuances of motion realism, precise spatio-temporal alignment, or textual ambiguity. Therefore, every candidate pair undergoes final manual verification by trained annotators using a custom interface that displays $\bm{M}_\text{src}$ and $\bm{M}_\text{tgt}$ side-by-side with visual highlights. Annotators perform a binary judgment, accepting a pair only if motions are physically plausible and the visual change exactly matches $L_{\text{edit}}^{\text{basic}}$.

To ensure high annotation quality and consistency, we implement a rigorous quality assurance protocol. This includes: 
1)~thorough annotator training and calibration using guidelines and canonical failure examples; 
2)~a production workflow with expert auditing, where a significant random sample of each annotator’s work is reviewed, and batches failing a quality threshold are re-annotated; 
and 3)~a final audit cycle. 
This process resulted in a high inter-annotator agreement and also led to the rejection or correction of candidates that passed automatic filtering, primarily due to subtle kinematic implausibilities or ambiguous instructions. The detailed protocol, including training steps, audit sampling rates, and interface specifications, is given in Appendix. This strict manual stage ensures the high fidelity of every triplet in the final dataset.

\begin{table*}[!t]
\begin{center}
\caption{
\textbf{Comparison of text-driven motion editing datasets.}
MotionFineEdit provides orders of magnitude more samples, supports fine-grained temporal and spatial control, and includes the first motion-grounded chain-of-thought annotations. 
\dag: Reports only countable paired motion with corrective descriptions. 
}
\label{table:dataset_comparison}
\setlength{\tabcolsep}{5pt} 
\begin{tabular}{lccccccc}
    \toprule[1pt]
    \multirow{3}{*}{Dataset} & \multicolumn{2}{c}{Motion Pair} & \multicolumn{3}{c}{Text} & \multirow{3}{*}{\makecell[c]{Easily \\ Scalable}} & \multirow{3}{*}{CoT} \\ 
    \cmidrule(r){2-3}\cmidrule(r){4-6}
     & Count & Intra-pair Length Flexibility & Count & Annotation Source & Temporal Cues  \\ 
    \midrule[0.5pt]
    MotionFix~\cite{athanasiou2024motionfix} $_\text{SIGGRAPH ASIA'24}$ & 6,730 & $\times$ & 6,730 & Human & $\times$ & $\times$ & $\times$ \\
    STANCE~\cite{jiang2025motionrefit} $_\text{CVPR'25}$ \dag & 4,411 & \checkmark & 4,411 & Human & $\times$ & $\times$ & $\times$ \\
    \textbf{MotionFineEdit (Ours)} & 35,003 & \checkmark & 981,162 & Auto $\rightarrow$ Human & \checkmark & \checkmark & \checkmark \\
    \bottomrule[1pt]
\end{tabular}
\end{center}
\end{table*}

\subsection{Enrichment and Composition}
\label{sec:enrichment}
The validated atomic triplets are fundamental but limited to single-step edits. To enable compositional reasoning and linguistic robustness, we enrich the dataset in two ways: by composing atomic edits into motion-grounded chain-of-thought annotations for multi-step reasoning, and by diversifying textual instructions via LLM rewriting. This transforms the atomic dataset into a richer resource for training models capable of complex, language-guided motion manipulation.

\subsubsection{Motion-Grounded CoT Annotation} 
\label{sec:cot}
A core contribution is the first \textit{motion-grounded chain-of-thought (CoT)} annotations, where each reasoning step is grounded in an aligned intermediate motion. This is enabled by the \textit{invertibility} of our atomic operations: for each edit with instruction $L_{\text{edit}}^{\text{basic}}$, we define an inverse $\text{Inv}(L_{\text{edit}}^{\text{basic}})$ that reverts it.  

Given validated atomic pairs $(\bm{M}_{\text{src}}, L_{\text{edit}_1}^{\text{basic}}, \bm{M}_{\text{tgt}}^1)$ and $(\bm{M}_{\text{src}}, L_{\text{edit}_2}^{\text{basic}}, \bm{M}_{\text{tgt}}^2)$ sharing the same source motion, we can compose them into a two-step complex edit:
\begin{align}
\bm{M}_{\text{tgt}}^1 \xrightarrow[]{\text{Inv}(L_{\text{edit}_1}^{\text{basic}})} \bm{M}_{\text{src}} \xrightarrow[]{L_{\text{edit}_2}^{\text{basic}}} \bm{M}_{\text{tgt}}^2,
\end{align} 
and form a complex editing pair:
\begin{equation}
    (\bm{M}_{\text{tgt}}^{1},  L_{\text{complex\_edit}}^{\text{basic}},  \bm{M}_{\text{tgt}}^{2}),
\end{equation}
where
\begin{equation}
    L_{\text{complex\_edit}}^{\text{basic}} = 
    [\text{Inv}(L_{\text{edit}_1}^{\text{basic}}) ;\ 
    L_{\text{edit}_2}^{\text{basic}} ].
\end{equation} 
The resulting annotation explicitly lists each step and its intermediate motion (see example in Fig.~\ref{fig:data_example} and Fig.~\ref{fig:intro}). This composition process can extend to longer chains, generating a large corpus of complex edits with verifiable reasoning traces. These motion-grounded CoT annotations provide essential supervision for the reasoning objective in RAGS. 

\subsubsection{Linguistic Diversification via LLM Rewriting}
\label{sec:llm_rewrite}
The basic instructions $L_{\text{edit}}^{\text{basic}}$ generated from templates, while precise, are stylistically rigid and may lead to model overfitting. To improve linguistic diversity and naturalness, we employ an LLM-based rewriting stage. For each atomic and complex edit instruction, we prompt Gemini~\cite{team2023gemini} to generate multiple paraphrases that preserve the original spatio-temporal semantics while varying phrasing, syntax, and lexical choice (\eg, changing ``raise the left hand'' to ``lift the left arm upwards''). This process significantly increases the linguistic variability of the dataset, encouraging the model to learn the underlying editing semantics rather than surface-level textual patterns.

\begin{figure}[!t]
    \centering
    \includegraphics[width=1.0\linewidth]{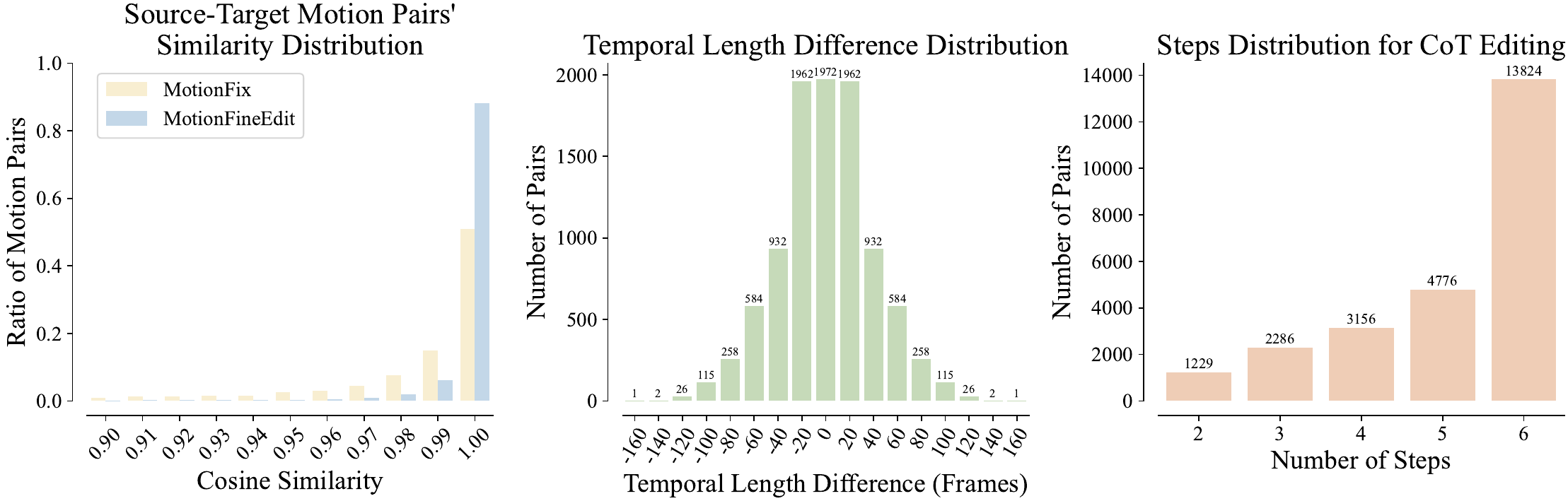}
    \caption{
    \textbf{Motion statistics of MotionFineEdit.} \textit{Left}:  Distribution of cosine similarity between source and target motions. MotionFineEdit pairs exhibit higher similarity than MotionFix, confirming more localized, fine-grained edits. 
    \textit{Middle}: Distribution of temporal length differences within pairs, demonstrating support for flexible duration changes. 
    \textit{Right}: Distribution of step counts in complex (CoT) edits, reflecting task complexity.
    }   
    \label{fig:motion_statistics}
\end{figure}

\subsection{Statistics Analysis}
\label{sec:statistics}
MotionFineEdit comprises \textbf{981K} fine-grained motion editing triplets, an order of magnitude larger than prior datasets (Tab.~\ref{table:dataset_comparison}). Critically, it includes \textbf{144K} complex triplets with motion-grounded CoT annotations, establishing the first large-scale benchmark for both fine-grained editing and reasoning.

To validate its utility, we analyze three key properties.
First, the \textbf{fine-grained nature} of edits is evidenced by significantly higher cosine similarity between source-target pairs in MotionFineEdit compared to MotionFix (Fig.~\ref{fig:motion_statistics}, left), confirming that edits are highly localized while preserving the global action context.
Second, the dataset supports substantial duration flexibility, with paired motions varying by up to 160 frames (Fig.~\ref{fig:motion_statistics}, middle), which is essential for realistic temporal control.
Third, to assess reasoning complexity, the distribution of step counts in CoT data shows a challenging long-tailed pattern (up to 6 steps, Fig.~\ref{fig:motion_statistics}, right), posing a rigorous test for compositional reasoning.
The dataset follows the similar data split as HumanML3D~\cite{humanml3d} to ensure fair comparison.

\section{Experimental Results}
\label{sec:experiments}

\subsection{Experimental Setup}
\label{sec:exp_setup}

\subsubsection{Datasets}
Empirical evaluations are conducted on four motion-text datasets.
\textbf{HumanML3D}~\cite{humanml3d} is a widely used motion-text paired dataset, comprising 14,616 motion sequences from AMASS~\cite{mahmood2019amass} and HumanAct12~\cite{humanact12}, along with 44,970 sequence-level textual captions.
\textbf{MotionFix}~\cite{athanasiou2024motionfix} is the first dataset for text-driven human motion editing, containing 6,730 triplets of source motions, target motions, and edit texts.
In both datasets, textual descriptions primarily capture overall motion semantics or holistic changes between motion pairs, making them well suited for coarse-grained motion-related tasks.
\textbf{FineMotion}~\cite{finemotion} re-labels all motions in HumanML3D with detailed body part movement annotations. Each motion is segmented into snippets at fixed temporal intervals, resulting in 420,968 snippets in total, where each snippet is paired with body part movement descriptions.
This dataset, together with \textbf{MotionFineEdit} proposed in this work, supports fine-grained motion-related tasks.

\subsubsection{Evaluation Metrics}
We adopt standard evaluation metrics.  
1)~For motion quality, realism, and text–motion alignment, we report Fréchet Inception Distance (FID), diversity, R-precision (Top-1/2/3), and multi-modal distance (MM-Dist)~\cite{t2mgpt, humanml3d}. For motion retrieval and similarity, we use recall (R@1/2/3), average recall (AvgR), and cosine similarity between motion embeddings.
2)~For generated text, we compute BLEU~\cite{papineni2002bleu}, ROUGE~\cite{rouge2004package}, and BERTScore~\cite{zhang2019bertscore} against references. To assess text–motion alignment, we also report R-precision and MM-Dist between the generated text and the corresponding motion.

\subsubsection{Implementation Details}
We follow the motion representation from HumanML3D~\cite{humanml3d} with dimensionality $d_m = 263$. The text-driven fine-grained motion generator $\mathcal{G}$ is fine-tuned using the configurations from MG-MotionLLM~\cite{mgmotionllm}. 
We instantiate MotionMERGE with three T5 backbones~\cite{t5}: small, base, and large. All models are optimized with AdamW~\cite{adamw}. The base model is pretrained for 500K iterations with a learning rate of $2 \times 10^{-4}$ and a batch size of 16, followed by task-specific instruction tuning for 300K iterations with a learning rate of $1 \times 10^{-4}$. The default hyperparameters for the loss function in Eq.~\eqref{eq:rags} are set as: $\lambda_\text{temp} = 2.0$, $\lambda_\text{local} = 2.0$, $\lambda_\text{cot} = 0.2$, and $\lambda_\text{coh}=1.0$, following uniform task sampling. Training was conducted on a single NVIDIA Tesla A100 (80GB) GPU. Additional details are provided in the Appendix.


\subsection{Comparisons on Text-Driven Motion Editing}
This task requires modifying a source motion $\bm{M}_\text{src}$ into a target motion $\bm{M}_\text{tgt}$ according to a textual instruction $L_\text{edit}$. Following standard practice~\cite{guo2025motionlab}, we report the average of five runs. Notably, prior diffusion-based methods require the target motion’s temporal length as an additional input, whereas MotionMERGE, as an autoregressive model, generates tokens until a stop condition, inferring length dynamically.

On the MotionFix dataset (Tab.~\ref{table:motionfix}), MotionMERGE achieves the best performance, outperforming the best prior method, MotionLab~\cite{guo2025motionlab}, by a clear margin (\eg, +1.8\% R@1, +3.2\% R@2) and significantly surpassing the specialized editor TMED~\cite{athanasiou2024motionfix}. This is noteworthy because it demonstrates the competitiveness of our unified autoregressive framework on a benchmark designed for specialized diffusion models. Furthermore, the gains on this coarse-grained dataset can be attributed to the \textit{richer motion-language representations} learned from fine-grained supervision during RAGS pre-training. This evidences that \textit{fine-grained modeling is fundamental for robust motion–language understanding}, even on coarse tasks. 
\begin{table}[!t]
\begin{center}
\caption{
    \textbf{Text-driven motion editing results on MotionFix.} $\dag$:~re-implemented with the HumanML3D representation for fair comparison. Best in \textbf{bold}, second-best \underline{underlined}. 
}
\label{table:motionfix}
\footnotesize
\setlength{\tabcolsep}{4.8pt} 
\begin{tabular}{lcccc}
    \toprule[1pt] 
    \multirow{2}{*}{Methods} & \multicolumn{4}{c}{generated-to-target retrieval} \\ 
    \cmidrule(r){2-5}
     & R@1 $\uparrow$ & R@2 $\uparrow$ & R@3 $\uparrow$ & AvgR $\downarrow$ \\
    \midrule[0.5pt]
    Real & 100.0 & 100.0 & 100.0 & 1.00 \\
    \midrule[0.5pt]
    TMED$\dag$~\cite{athanasiou2024motionfix} $_{\text{SIGGRAPH Asia'24}}$ & 35.16 & 51.43 & 60.81 & 5.11  \\
    
    MotionLab~\cite{guo2025motionlab} $_{\text{ICCV'25}}$ & \underline{56.34} & \underline{70.40} & \underline{77.24} & \underline{3.54} \\
    \textbf{MotionMERGE (Ours)} & \textbf{58.15} & \textbf{73.55} & \textbf{80.77} & \textbf{2.85} \\
    \bottomrule[1pt]
\end{tabular}
\end{center}
\end{table}

\begin{table}[!t]
\begin{center}
\caption{
    \textbf{Text-driven Fine-grained Motion Editing Results on MotionFineEdit}.
    $\dag$:~re-implemented with the HumanML3D representation for fair comparison.
    Best in \textbf{bold}, second-best \underline{underlined}.
}
\label{table:MotionFineEdit}
\footnotesize
\setlength{\tabcolsep}{0.8pt} 
\begin{tabular}{lccccccc}
    \toprule[1pt]
    \multirow{2}{*}{Methods} & \multicolumn{3}{c}{Sequence G2T retrieval} & \multicolumn{3}{c}{Snippet G2T retrieval} \\ 
    \cmidrule(r){2-4}\cmidrule(r){5-7}
     & R@1 $\uparrow$ & R@3 $\uparrow$ & AvgR $\downarrow$ & R@1 $\uparrow$ & R@3 $\uparrow$ & AvgR $\downarrow$ \\
    \midrule[0.5pt]
    Real & 100.0 & 100.0 & 1.00 & 100.0 & 100.0 & 1.00 \\
    \midrule[0.5pt]
    \textcolor{gray}{\textit{Atomic}} \\
    TMED$\dag$~\cite{athanasiou2024motionfix} $_{\text{SIGGRAPH Asia'24}}$ & \underline{9.32} & \underline{22.50} & \underline{11.42} & \underline{7.16} & \underline{17.56} & \underline{11.72}  \\
    MotionLab~\cite{guo2025motionlab} $_{\text{ICCV'25}}$ & 6.80 & 19.90 & 12.61 & 6.33 & 16.24 & 13.97 \\    
    \textbf{MotionMERGE (Ours)} & \textbf{79.84} & \textbf{93.84} & \textbf{1.56} & \textbf{41.81} & \textbf{52.03} & \textbf{6.63}  \\
    \midrule[0.5pt]
    \textcolor{gray}{\textit{Complex}} \\
    TMED$\dag$~\cite{athanasiou2024motionfix} $_{\text{SIGGRAPH Asia'24}}$ & \underline{6.13} & \underline{14.15} & 15.31 & \underline{4.11} & \underline{11.31} & \underline{14.55} \\
    MotionLab~\cite{guo2025motionlab} $_{\text{ICCV'25}}$ & 2.42 & 9.56 & \underline{15.02} & 2.79 & 9.17 & 15.61 \\
    \textbf{MotionMERGE (Ours)} & \textbf{20.85} & \textbf{35.30} & \textbf{9.12} & \textbf{11.39} & \textbf{25.98} & \textbf{10.44} \\
    \bottomrule[1pt]
\end{tabular}
\end{center}
\end{table}

Then, we evaluate the results of fine-grained editing on MotionFineEdit (Tab.~\ref{table:MotionFineEdit}) using both sequence‑level and snippet‑level retrieval, with the latter measuring alignment within localized 10-frame intervals. The results reveal a stark granularity gap: prior methods perform near random chance, especially under snippet-level scrutiny, proving they lack mechanisms for precise spatio-temporal grounding. In contrast, MotionMERGE achieves superior performance, a direct result of its synergistic design: RAGS pre-training instills necessary biases (\eg, localization and temporal priors) and training on MotionFineEdit provides essential supervision. The model maintains a strong lead in snippet-level metrics (\eg, 41.81\% vs. 7.16\% R@1 on atomic edits), proving its edits are locally accurate. Its advantage is even more pronounced on complex edits, benefiting from the motion-grounded chain-of-thought supervision in RAGS.

These results yield two important insights. 1)~The performance reversal, in which the specialized editor TMED surpasses the unified framework MotionLab, exposes a fundamental \textit{specialization–generalization trade-off} in fine-grained editing. Our work shows that a unified framework can overcome this by being tailored for multiple granularities. 2)~The substantial drop from sequence-level to snippet-level metrics (\eg, MotionMERGE’s R@1 falls from 79.84\% to 41.81\%) demonstrates that \textit{standard holistic evaluation is insufficient} and can mask severe local misalignments, advocating for more discriminative, localized evaluation protocols. Together, these insights validate that advancing fine-grained motion editing requires \textit{co-evolving data, model, and training paradigm}.  

\begin{figure*}[!t]
\begin{center}
\includegraphics[width=1.0\linewidth]{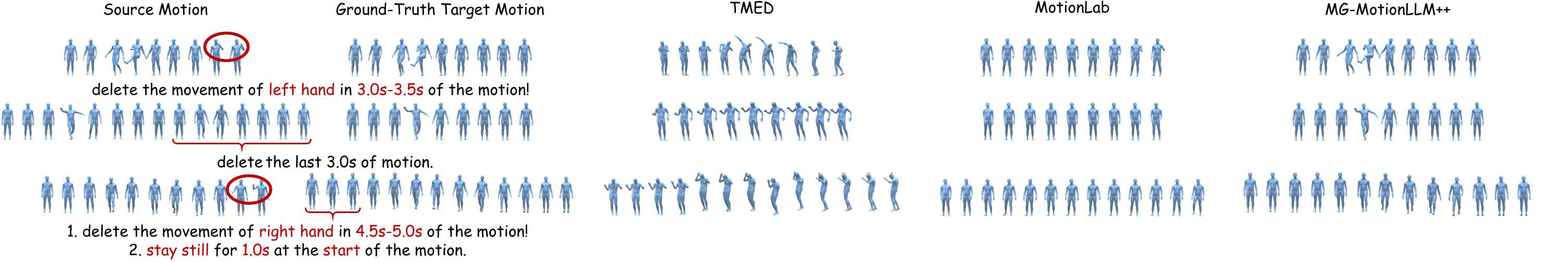}
\end{center}
\caption{
\textbf{Qualitative results of text-driven fine-grained human motion editing on MotionFineEdit.} 
Rows illustrate atomic (top: spatial, middle: temporal) and complex (bottom: combined) editing tasks. Motions are sampled every 0.5 seconds. 
}
\label{fig:edit_results}
\end{figure*}

Fig.~\ref{fig:edit_results} provides qualitative validation, visually confirming MotionMERGE’s precise execution of fine-grained edits per instruction, in contrast to the artifacts or failures common in baseline methods. This aligns with and reinforces our quantitative findings. 

\subsection{Comparisons on Motion Captioning}
\label{sec:results_captioning}
To assess motion understanding, we evaluate the task of generating textual descriptions for given motions. 
Following~\cite{openmotionlab_motiongpt}, we use the original ground-truth texts for precise assessment, avoiding preprocessing that may discard linguistic nuance.

As shown in Tab.~\ref{table:humanml3d_m2t}, MotionMERGE achieves state-of-the-art performance of generating coarse textual descriptions on HumanML3D, with leading scores on R-Precision (Top-1: 0.602), MM-Dist (2.510), and BERTScore (38.8). This demonstrates that our fine-grained pre-training enhances semantic alignment even for coarse-grained generation.
Interestingly, prior unified models (LaMP-M2T, MotionGPT) achieve higher scores on lexical overlap metrics like Bleu@4 (13.04 and 12.47). We attribute this to a fundamental trade-off in training objective: standard training optimizes for dataset-style fluency (mimicking the distribution of coarse captions), while RAGS pre-training optimizes for fine-grained semantic grounding. The latter enriches the model's descriptive vocabulary at the cost of lower lexical conformity to the test set, but yields stronger motion–text correspondence.

\begin{table}[!t]
\begin{center}
\caption{
\textbf{Motion captioning results on HumanML3D~\cite{humanml3d}.} Best in \textbf{bold}, second-best \underline{underlined}. *: from~\cite{openmotionlab_motiongpt}. \textbf{BS}: BERTScore. 
}
\label{table:humanml3d_m2t}
\footnotesize
\setlength{\tabcolsep}{1.8pt} 
\begin{tabular}{lcccccc}
    \toprule[1pt]
    \multirow{2}{*}{Methods} & \multicolumn{2}{c}{R-Precision $\uparrow$} & \multirow{2}{*}{MM-Dist $\downarrow$}  & \multirow{2}{*}{Bleu@4 $\uparrow$} & \multirow{2}{*}{BS $\uparrow$} \\ 
    \cmidrule(r){2-3}
     & Top-1 & Top-3 &  \\
    \midrule[0.5pt]
    Real & 0.523 & 0.828 & 2.901 & - & - \\
    \midrule[0.5pt]
    TM2T*~\cite{guo2022tm2t} $_{\text{ECCV'22}}$ & 0.516 & 0.823 & 2.935 & 7.00 & 32.2 \\
    MotionGPT~\cite{openmotionlab_motiongpt} $_{\text{NeurIPS'23}}$ & 0.543 & 0.827 & 2.821 & \underline{12.47} & 32.4 \\
    LaMP-M2T~\cite{lamp} $_{\text{ICLR'25}}$ & 0.547 & 0.831 & 2.808 & \textbf{13.04} & 32.7 \\
    MG-MotionLLM~\cite{mgmotionllm} $_{\text{CVPR'25}}$ & \underline{0.592} & \underline{0.866} & \underline{2.581} & 8.06 & \underline{36.7}  \\
    \textbf{MotionMERGE (Ours)} & \textbf{0.602} & \textbf{0.880} & \textbf{2.510} & 9.44 & \textbf{38.8}   \\
    \bottomrule[1pt]
\end{tabular}
\end{center}
\end{table}

Beyond coarse captioning, we evaluate the generation of detailed descriptions on the FineMotion dataset~\cite{finemotion}, where the task is to describe body-part movements per 10-frame snippet. MotionMERGE consistently outperforms its predecessor MG-MotionLLM~\cite{mgmotionllm} across all model sizes and metrics, with performance scaling with capacity (Tab.~\ref{table:m2dt}). Notably, it achieves higher absolute scores on this challenging task (\eg, BERTScore up to 52.3) than on coarse captioning (\eg, BERTScore 38.8 on HumanML3D). This counter-intuitive result reveals a key property of fine-grained data: snippet-level descriptions are more structured and templatized (\eg, ``move right arm forward''), providing a constrained, consistent generation target compared to the open-ended diversity of coarse descriptions. MotionMERGE’s superior performance stems from its RAGS pre-training, which explicitly optimizes for aligning partial motion segments with localized textual descriptions. Together, these results demonstrate that our approach effectively learns the structured, local motion–language correspondences required for fine-grained description.  
\begin{table}[!t]
\begin{center}
\caption{
    \textbf{Detailed Motion Captioning} results on the FineMotion~\cite{finemotion} test set.
    \textbf{bold}, second-best \underline{underlined}.
}
\label{table:m2dt}
\footnotesize
\setlength{\tabcolsep}{1.8pt} 
\begin{tabular}{lccccc}
    \toprule[1pt]
     Methods & Bleu@1 $\uparrow$ & Bleu@4 $\uparrow$  & Bleu@7 $\uparrow$ & Rouge $\uparrow$ & BertScore $\uparrow$ \\
    \midrule[0.5pt]
    \multicolumn{6}{l}{\textcolor{gray}{\textit{Model Size: \textbf{Small}}}} \\
    MG-MotionLLM & 65.33 & 44.93 & 31.99 & 57.8 & 47.4 \\
    MotionMERGE & 66.55 & 46.72 & 34.14 & 59.4 & 49.3  \\
    \midrule[0.5pt]
    \multicolumn{6}{l}{\textcolor{gray}{\textit{Model Size: \textbf{Base}}}} \\
    MG-MotionLLM & 66.57 & 46.94 & 34.43 & 59.6 & 49.8 \\
    MotionMERGE & 67.15 & 47.69 & 35.31 & 60.4 & 50.7 \\
    \midrule[0.5pt]
    \multicolumn{6}{l}{\textcolor{gray}{\textit{Model Size: \textbf{Large}}}} \\
    MG-MotionLLM & \textbf{68.25} & \underline{49.26} & \underline{37.07} & \underline{61.4} & \underline{52.2} \\
    MotionMERGE & \underline{68.19} & \textbf{49.40} & \textbf{37.37} & \textbf{61.6} & \textbf{52.3} \\
    \bottomrule[1pt]
\end{tabular}
\end{center}
\end{table}

\subsection{Comparisons on Text-Driven Motion Generation}
\label{sec:results_generation}

This task involves generating motions based on textual descriptions. 
All results are reported with a 95\% confidence interval, obtained from 20 repeated trials. 
We evaluate both \textit{specialized motion generators} and \textit{unified frameworks}.
\begin{table}[!t]
\caption{
\textbf{Text-to-motion generation results on HumanML3D.} Methods are grouped by specialization: dedicated motion generation (top block) vs. unified models (bottom block). Best in each block in \textbf{bold}.
} \label{table:humanml3d_t2m}
\begin{center}
\footnotesize
\setlength{\tabcolsep}{2.8pt}
\resizebox{\linewidth}{!}{
\begin{tabular}{lcccc}
    \toprule[1pt]
    Methods & R-Top3 $\uparrow$ & FID $\downarrow$ & MM-Dist $\downarrow$  & Diversity $\uparrow$ \\ 
    \midrule[0.5pt]
    Real motion & 0.797$^{\pm.002}$ & 0.002$^{\pm.000}$ & 2.974$^{\pm.008}$ & 9.503$^{\pm.065}$ \\
    \midrule[0.5pt]
    \multicolumn{5}{l}{\textcolor{gray}{\textit{Motion Generation Only}}}
     \\
    TEMOS~\cite{petrovich2022temos} $_{\text{ECCV'22}}$ & 0.722$^{\pm.002}$ & 3.734$^{\pm.028}$ & 3.703$^{\pm.008}$ & 8.973$^{\pm.071}$ \\
    Guo et al.\cite{humanml3d} $_{\text{CVPR'22}}$ & 0.736$^{\pm.002}$ & 1.087$^{\pm.021}$ & 3.347$^{\pm.008}$ & 9.175$^{\pm.083}$ \\
    MDM~\cite{mdm} $_{\text{ICLR'23}}$ & 0.749$^{\pm.006}$ & 0.489$^{\pm.047}$ & 3.330$^{\pm.025}$ & 9.920$^{\pm.083}$ \\
    T2M-GPT~\cite{t2mgpt} $_{\text{CVPR'23}}$ & 0.775$^{\pm.002}$ & 0.141$^{\pm.004}$ & 3.121$^{\pm.009}$ & \textbf{9.722}$^{\pm.082}$ \\
    MotionDiffuse~\cite{zhang2024motiondiffuse} $_{\text{TPAMI'24}}$ & 0.782$^{\pm.001}$ & 0.630$^{\pm.001}$ & 3.113$^{\pm.001}$ & 9.410$^{\pm.049}$ \\
    FineMoGen~\cite{zhang2023finemogen} $_{\text{NeurIPS'23}}$ & 0.784$^{\pm.002}$ & 0.151$^{\pm.008}$ & 2.998$^{\pm.008}$ & 9.263$^{\pm.094}$ \\
    MoMask~\cite{guo2024momask} $_{\text{CVPR'24}}$ & 0.807$^{\pm.002}$ & 0.045$^{\pm.002}$ & 2.958$^{\pm.008}$ & - \\
    EnergyMoGen~\cite{zhang2025energymogen} $_{\text{CVPR'25}}$ & 0.815$^{\pm.002}$ & 0.176$^{\pm.006}$ & 2.931$^{\pm.007}$ & 9.500$^{\pm.091}$ \\
    KinMo~\cite{zhang2025kinmo} $_{\text{ICCV'25}}$ & 0.821$^{\pm.003}$ & 0.039$^{\pm.003}$ & 2.901$^{\pm.010}$ & 9.674$^{\pm.058}$ \\
    LaMP-T2M~\cite{lamp} $_{\text{ICLR'25}}$ & \textbf{0.843}$^{\pm.001}$ & \textbf{0.032}$^{\pm.002}$ & \textbf{2.759}$^{\pm.007}$ & 9.571$^{\pm.069}$ \\
    \midrule[0.5pt]
    \multicolumn{5}{l}{\textcolor{gray}{\textit{Unified Motion Modeling}}}     \\
    MotionGPT~\cite{openmotionlab_motiongpt} $_{\text{NeurIPS'23}}$ & 0.778$^{\pm.002}$ & 0.232$^{\pm.008}$ & 3.096$^{\pm.008}$ & 9.528$^{\pm.071}$ \\
    MG-MotionLLM~\cite{mgmotionllm} $_{\text{CVPR'25}}$ & 0.802$^{\pm.003}$ & 0.303$^{\pm.010}$ & 2.952$^{\pm.009}$ & 9.960$^{\pm.073}$ \\
    MotionLab~\cite{guo2025motionlab} $_{\text{ICCV'25}}$ & 0.810\hspace{2.5em} & \textbf{0.167}\hspace{2.5em} & 2.830\hspace{2.5em} & 9.593\hspace{2.5em} \\
    \textbf{MotionMERGE (Ours)} &  \textbf{0.826}$^{\pm.002}$ & 0.233$^{\pm.005}$ & \textbf{2.818}$^{\pm.006}$ & \textbf{9.970}$^{\pm.116}$ \\
    \bottomrule[1pt]
\end{tabular}
}
\end{center}
\end{table}

\begin{figure*}[!b]
\begin{center}
\includegraphics[width=0.9\linewidth]{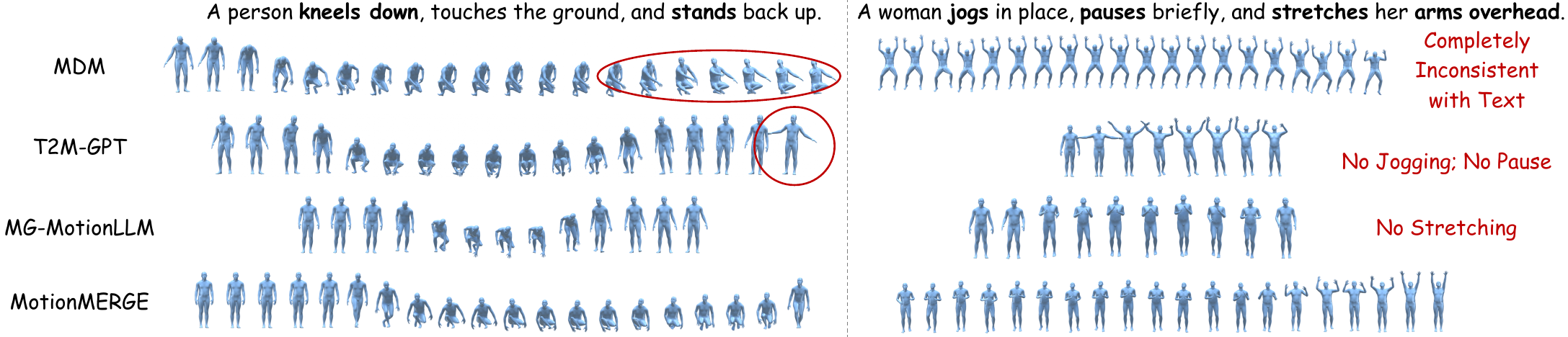}
\end{center}
\caption{
\textbf{Qualitative text-to-motion results}. MotionMERGE generates motions that accurately match textual descriptions, including complex multi-action sequences.
}
\label{fig:t2m_gen_qualitative_results}
\end{figure*}

Tab.~\ref{table:humanml3d_t2m} summarizes the comparison results on HumanML3D for generating motions from coarse textual descriptions. It shows that MotionMERGE achieves the best performance among unified frameworks, leading on R-Top3 (0.826), MM-Dist (2.818), and Diversity (9.970), and notably surpasses its predecessor MG-MotionLLM~\cite{mgmotionllm}. While it does not surpass the state-of-the-art specialized model LaMP-T2M~\cite{lamp}, the gap is remarkably small, especially on motion quality (MM-Dist: 2.818 vs. 2.759) and diversity. This indicates that the performance penalty for a unified, multi-task architecture is minimal, a key outcome of our design. The RAGS pre-training instills a richer, more structured motion prior by forcing fine-grained, temporally-localized alignment, which benefits coarse generation. Concurrently, the high-quality, varied motions in MotionFineEdit act as effective data augmentation, enriching the model’s motion repertoire. Together, these advances validate a critical insight: investing in fine-grained understanding and data quality is a viable path to building high-performance unified models, mitigating the traditional trade-off between specialization and generality. Qualitative results (Fig.~\ref{fig:t2m_gen_qualitative_results}) confirm that MotionMERGE accurately follows complex, multi-action descriptions.

Furthermore, we introduce a snippet-level evaluation to directly assess fine-grained temporal alignment, a prerequisite for precise controllable generation. It computes retrieval accuracy (R-precision) and cosine similarity between corresponding temporal snippets of generated and ground-truth motions (Tab.~\ref{table:tdt2m}), measuring alignment with the target’s temporal structure beyond holistic similarity. 
\begin{table}[!t]
\centering
\caption{
\textbf{Snippet-level temporal alignment of generated motions.} Results grouped by input granularity (coarse/fine). Best in each block in \textbf{bold}. 
}
\label{table:tdt2m}
\footnotesize
\setlength{\tabcolsep}{2pt}
\begin{tabular}{lccccc}
    \toprule[1pt]
    \multirow{2}{*}{Methods} & \multicolumn{3}{c}{R-Precision $\uparrow$} & \multirow{2}{*}{AvgR $\downarrow$} & \multirow{2}{*}{Cos. Sim. $\uparrow$} \\ 
    \cmidrule(r){2-4}
     & Top-1 & Top-2 & Top-3 &  \\
    \midrule[0.5pt]
    \multicolumn{5}{l}{\textcolor{gray}{\textit{Coarse-grained}}}
     \\
    MDM~\cite{mdm} $_{\text{ICLR'23}}$ & 13.79 & 21.69 & 27.56 & 11.24 & 0.779 \\
    T2M-GPT~\cite{t2mgpt} $_{\text{CVPR'23}}$ & 13.82 & 20.56 & 25.58 & 11.00 & 0.845 \\
    MoMask~\cite{guo2024momask} $_{\text{CVPR'24}}$ & 12.93 & 20.32 & 25.91 & 10.11 & 0.830 \\
    MG-MotionLLM~\cite{mgmotionllm}  $_{\text{CVPR'25}}$ & 15.48 & 22.92 & 28.35 & 10.35 & 0.850 \\
    \textbf{MotionMERGE (Ours)} & \textbf{16.43} & \textbf{24.14} & \textbf{29.71} & \textbf{10.03} & \textbf{0.855} \\
    \midrule[0.5pt]
    \multicolumn{5}{l}{\textcolor{gray}{\textit{Fine-grained}}}
     \\
    (T\&DT)-MDM~\cite{mdm}  $_{\text{ICLR'23}}$ \dag & 15.25 & 23.53 & 29.53 & 10.82 & 0.794  \\
    (T\&DT)2M-GPT~\cite{t2mgpt} $_{\text{CVPR'23}}$ \dag & 17.69 & 25.98 & 32.03 & 9.23 & 0.874  \\
    (T\&DT)-MoMask~\cite{guo2024momask} $_{\text{CVPR'24}}$ \dag & 24.60 & 34.33 & 40.52 & 7.73 & 0.884 \\
    MG-MotionLLM~\cite{mgmotionllm}  $_{\text{CVPR'25}}$ & 37.00 & 47.74 & 54.39 & 5.06 & 0.938 \\
    \textbf{MotionMERGE (Ours)} & \textbf{38.13} & \textbf{48.78} & \textbf{55.36} & \textbf{4.92} & \textbf{0.942} \\
    \bottomrule[1pt]
\end{tabular}
\end{table}

The results reveal two clear patterns. 
1)~Fine-grained textual conditioning yields substantially higher snippet-level alignment (\eg, 38.13\% vs. 16.43\% R@1 for MotionMERGE) than coarse conditioning across all methods. This confirms that detailed descriptions provide a stronger temporal scaffold. 
2)~MotionMERGE achieves the best alignment under both conditioning types, notably surpassing its predecessor MG-MotionLLM (\eg, 38.13\% vs. 37.00\% R@1 under fine-grained conditioning) and outperforming re-implemented specialists like (T\&DT)-MoMask. This fidelity is a direct outcome of the temporal grounding objective in our RAGS pre-training, which forces the model to learn explicit associations between partial fine-grained textual descriptions and specific time intervals. Thus, high snippet-level similarity provides orthogonal validation that our framework achieves not only semantic but also temporal precision.

\begin{figure*}[!t]
\begin{center}
\includegraphics[width=1.0\linewidth]{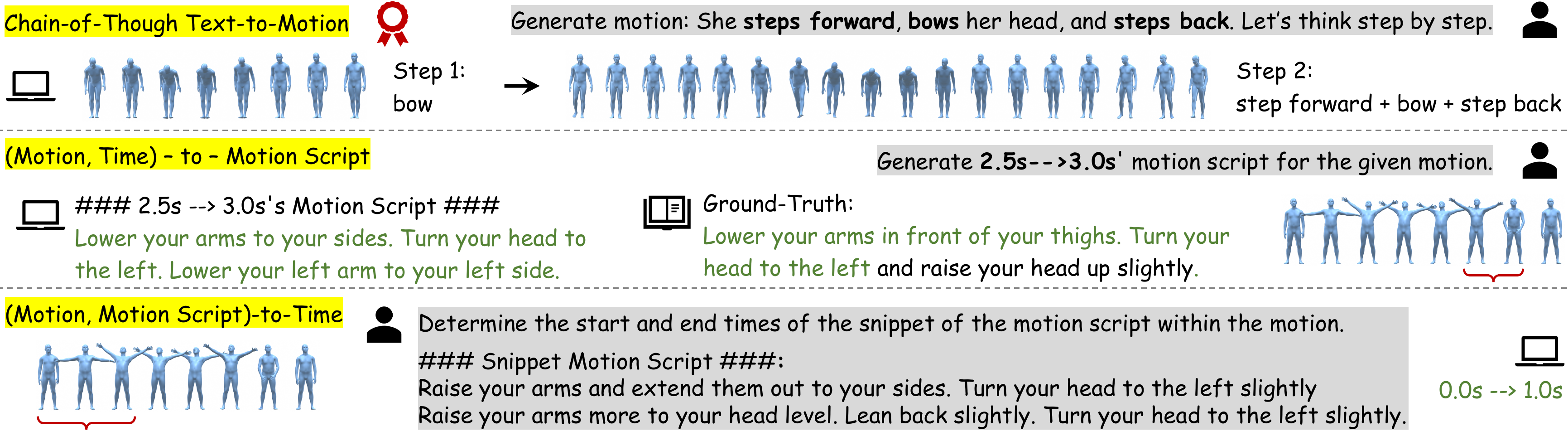}
\end{center}
\caption{
\textbf{Novel motion-language tasks.} Examples include chain-of-thought motion generation (\textit{top}), a zero-shot emergent reasoning capability, fine-grained captioning of partial sequences (\textit{middle}), and motion localization via textual description (\textit{bottom}). 
}
\label{fig:other_app_examples}
\end{figure*}

\subsection{More Results of Novel Applications.} 
Our RAGS pre-training enables \textbf{zero-shot} transfer of structured reasoning to novel, complex motion-language tasks.
As shown in the top part of Fig.~\ref{fig:other_app_examples}, the model provides direct evidence of the structured reasoning instilled by RAGS. In chain-of-thought generation, the model decomposes the instruction step forward, bow, step back and strategically generates the middle action (bow) first. This non-chronological order suggests an internal planning process that identifies and establishes a pivotal anchor to scaffold the full narrative. 
Crucially, these capabilities are elicited zero-shot via simple prompts, confirming that the reasoning skills are \textit{generalizable} properties of the model, not task-specific fittings.

Meanwhile, Fig.~\ref{fig:other_app_examples} also showcases the model's performance on novel tasks, such as \textit{partial-sequence captioning} and \textit{textual motion localization}, executed via task-specific prompting without further fine-tuning.
For motion localization, the precise mapping of a detailed textual snippet to a tight temporal window (0.0s–1.0s) demonstrates an ability to \textit{parse and ground} linguistic descriptions of intent and duration. Similarly, fine-grained interval captioning confirms a \textit{segmented understanding of motion}.

\begin{table*}[!b]
\begin{center}
\caption{
    \textbf{Ablation of training strategy.} Best in \textbf{bold}, second-best \underline{underlined}. 
    \textbf{T2M}: Text-Driven Motion Generation;
    \textbf{(T+DT)2M}: Text-Driven Fine-Grained Motion Generation;
    \textbf{M2T}: Motion Captioning;
    \textbf{M2DT}: Motion Detailed Captioning;
    \textbf{ME}: Text-Driven Motion Editing;
    \textbf{FG-ME}: Text-Driven Fine-Grained Motion Editing.
}
\label{table:ablation_training_strategy}
\footnotesize
\setlength{\tabcolsep}{1.8pt} 
\begin{tabular}{ccccccccccccccc}
    \toprule[1pt]
    \multirow{2}{*}{\makecell{Row \\ No.}} & \multirow{2}{*}{\makecell{Pretraining \\ Granularity}} & \multirow{2}{*}{Ins.} & \multicolumn{2}{c}{T2M} & \multicolumn{2}{c}{(T+DT)2M} & \multicolumn{2}{c}{M2T}  & \multicolumn{2}{c}{M2DT} & \multicolumn{2}{c}{ME}  & \multicolumn{2}{c}{FG-ME} \\ 
    \cmidrule(r){4-5}\cmidrule(r){6-7}\cmidrule(r){8-9}\cmidrule(r){10-11}\cmidrule(r){12-13}\cmidrule(r){14-15}
     & & & Top-3 $\uparrow$ & FID $\downarrow$ & Top-3 $\uparrow$ & Cos. Sim. $\uparrow$ & Top-1 $\uparrow$ & BertScore $\uparrow$ & Bleu@4 $\uparrow$ & BertScore $\uparrow$ & R@3 $\uparrow$ & AvgR $\downarrow$ & Seq. Avg $\downarrow$ & Sni. Avg $\downarrow$ \\
    \midrule[0.5pt]
    \multicolumn{3}{c}{\textcolor{gray}{Real}} & 0.797 & 0.002 & 100.00 & 1.000 & 0.523 & - & - & - & 100.0 & 1.00 & 1.00 & 1.00 \\
    \midrule[0.5pt]
    (1) & - & \checkmark & 0.773 & 0.485 & \underline{54.39} & \underline{0.938} & 0.516 & 22.7 & 42.18 & 44.6 & 73.95 & 3.78 & 1.60 & 7.42 \\
    (2) & Coarse & - &  0.778 & 0.496 & - & - & 0.506 & \textbf{39.0} & - & - & 74.19 & 3.73 & - & - \\
    (3) & Fine & - & - & - & 53.13 & 0.937 & - & - & 42.96 & 45.1 & - & - & 1.58 & 6.88 \\
    (4) & Coarse + Fine & - & \underline{0.806} & \underline{0.331} & 52.38 & 0.934 & \underline{0.533} & \textbf{39.0} & \underline{46.43} & \underline{49.0} & \underline{74.62} & \underline{3.62} & \textbf{1.54} & \underline{6.65} \\
    (5) & Coarse + Fine & \checkmark & \textbf{0.826} & \textbf{0.233} & \textbf{55.36} & \textbf{0.942} & \textbf{0.602} & 38.8 & \textbf{47.69} & \textbf{50.7} & \textbf{80.77} & \textbf{2.85} & \underline{1.56} &  \textbf{6.63} \\
    \bottomrule[1pt]
\end{tabular}
\end{center}
\end{table*}

These results qualitatively validate that MotionMERGE, via RAGS pre-training, moves beyond reactive pattern completion. It demonstrates an emergent capacity for decomposition, temporal inference, and structured synthesis, core facets of reasoning that are essential for \textit{interpretable and controllable} motion modeling. While comprehensive quantitative benchmarking of such reasoning remains an open challenge, the behaviors shown here constitute a compelling proof-of-concept that our approach successfully integrates reasoning into a unified motion-language framework.


\subsection{Ablation Study}
\label{sec:ablation}
In this subsection, we comprehensively evaluate the impact of different designs in MotionMERGE.

\subsubsection{Ablation of Training Strategy}
\label{sec:ablation_training}
We ablate the proposed RAGS training strategy in Tab.~\ref{table:ablation_training_strategy}. The results yield three principal findings.
1)~The RAGS pre-trained model (Row 4) consistently outperforms the baseline that is directly instruction-tuned on most tasks (Row 1), despite using only about 1/30 the per-task iterations. This demonstrates that RAGS pre-training is a highly data- and compute-efficient pathway to a generalizable motion-language foundation.
2)~Models pre-trained solely on coarse (Row 2) or fine (Row 3) tasks exhibit strong but narrow expertise. In contrast, the RAGS model (Row 4), trained on both granularities, achieves the best or competitive performance on both types of tasks. This indicates a genuine synergistic interaction: fine-grained learning appears to refine coarse semantic understanding, while coarse exposure provides a richer context for fine-grained tasks, validating the core design of RAGS.
3)~Final instruction tuning (Row 5) yields the best overall performance, confirming that task-specific adaptation fully unlocks the potential of the pre-trained foundation.

\begin{table*}[!t]
\caption{
    \textbf{Ablation of model sizes.} Best in \textbf{bold}, second-best \underline{underlined}. 
    \textbf{T2M}: Text-Driven Motion Generation;
    \textbf{(T+DT)2M}: Text-Driven Fine-Grained Motion Generation;
    \textbf{M2T}: Motion Captioning;
    \textbf{M2DT}: Motion Detailed Captioning;
    \textbf{ME}: Text-Driven Motion Editing;
    \textbf{FG-ME}: Text-Driven Fine-Grained Motion Editing.
}
\vspace{-15pt}
\label{table:ablation_model_size}
\begin{center}
\footnotesize
\setlength{\tabcolsep}{3.8pt} 
\begin{tabular}{lccccccccccccc}
    \toprule[1pt]
    \multirow{2}{*}{Size} & \multirow{2}{*}{Param.} & \multicolumn{2}{c}{T2M} & \multicolumn{2}{c}{(T+DT)2M} & \multicolumn{2}{c}{M2T}  & \multicolumn{2}{c}{M2DT} & \multicolumn{2}{c}{ME}  & \multicolumn{2}{c}{FG-ME} \\ 
    \cmidrule(r){3-4}\cmidrule(r){5-6}\cmidrule(r){7-8}\cmidrule(r){9-10}\cmidrule(r){11-12}\cmidrule(r){13-14}
    &  & Top-3 $\uparrow$ & FID $\downarrow$ & Top-3 $\uparrow$ & Cos. Sim. $\uparrow$ & Top-1 $\uparrow$ & BertScore $\uparrow$ & Bleu@4 $\uparrow$ & BertScore $\uparrow$ & R@3 $\uparrow$ & AvgR $\downarrow$ & Seq. Avg $\downarrow$ & Sni. Avg $\downarrow$ \\
    \midrule[0.5pt]
    \multicolumn{2}{c}{\textcolor{gray}{Real}} & 0.797 & 0.002 & 100.00 & 1.000 & 0.523 & - & - & - & 100.0 & 1.00 & 1.00 & 1.00 \\
    \midrule[0.5pt]
    Small & 60M & \underline{0.806} & 0.352 & \underline{53.52} & \underline{0.937} & 0.593 & \textbf{40.0} & 46.72 & 49.3 & \underline{78.33} & \underline{3.15} & \underline{1.59} & 7.19 \\
    Base  & 220M & \textbf{0.826} & \textbf{0.233} & \textbf{55.36} & \textbf{0.942} & \textbf{0.602} & 38.8 & 47.69 & 50.7 & \textbf{80.77} & \textbf{2.85} & \textbf{1.56} & \textbf{6.63} \\
    Large & 770M & 0.799 & \underline{0.339} & 53.20 & \underline{0.937} & \underline{0.596} & \underline{39.1} & \textbf{49.40} & \textbf{52.3} & 78.19 & \underline{3.15} & 1.69 & \underline{6.70} \\
    \bottomrule[1pt]
\end{tabular}
\end{center}
\end{table*}

\begin{figure}[!t]
    \centering
    \includegraphics[width=1.0\linewidth]{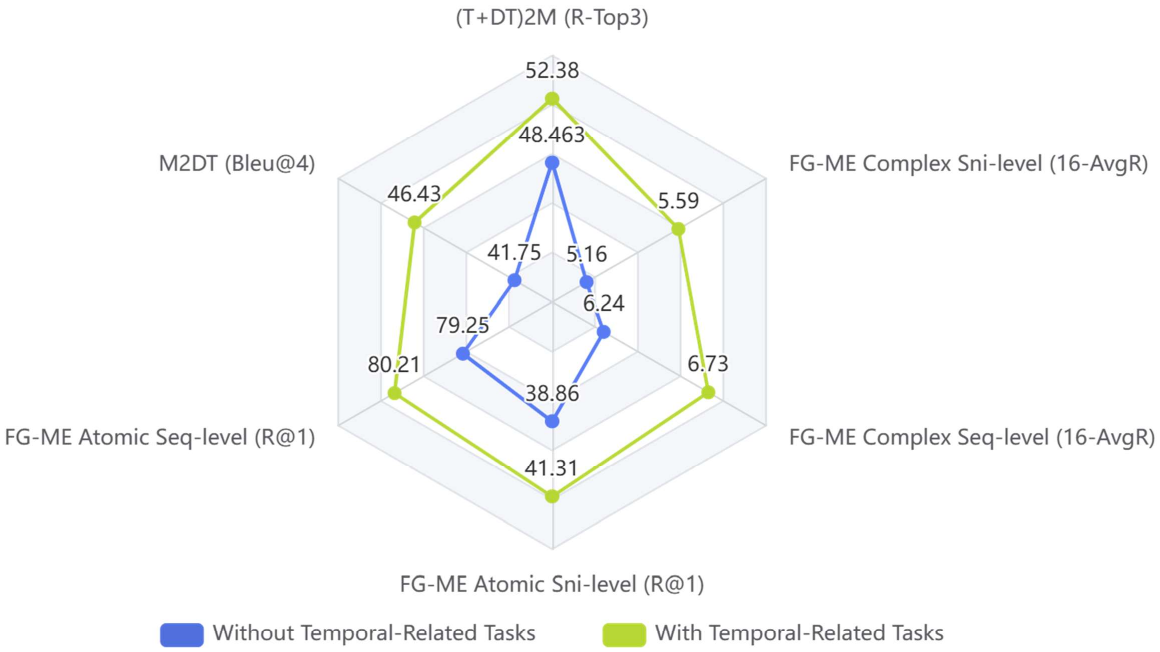}
    \caption{
    \textbf{Ablation of temporal-aware pre-training on fine-grained tasks}. (T+DT)2M: fine-grained text-to-motion; M2DT: detailed motion captioning; FG-ME: fine-grained motion editing. Higher is better for all metrics; (16-AvgR) inverts AvgR (random baseline: 16). 
    }   
    \label{fig:fig_compaison_with_without_temp}
\end{figure}

\subsubsection{Ablation of Model Size}
\label{sec:ablation_size}
We ablate the model size of MotionMERGE in Tab.~\ref{table:ablation_model_size}. The results reveal three principal scaling trends.  
1)~Scaling from Small (60M) to Base (220M) parameters yields substantial, near-universal gains, establishing the 220M scale as a \textit{performance-compute Pareto optimum} for the current data regime. 2)~Further scaling to Large (770M) leads to diminishing returns and even degradation on core tasks like T2M (FID: 0.339 vs. 0.233) and fine-grained editing, aligning with observations in MotionGPT~\cite{openmotionlab_motiongpt} and indicating potential overfitting tendencies when model capacity vastly exceeds the dataset’s scale and diversity. 
3)~Scaling impact is task-dependent: generation tasks (T2M, (T+DT)2M) are more sensitive to over-parameterization, likely due to their greater susceptibility to memorizing data artifacts, while understanding tasks (M2T, M2DT) remain more stable. This differential sensitivity indicates that one-size-fits-all scaling laws do not apply.

\subsubsection{Importance of Temporal Awareness}
\label{sec:ablation_temporal}
To evaluate the necessity of explicit temporal modeling, we ablate the temporal-aware tasks from the RAGS pre-training stage. We compare the full RAGS model against a variant pre-trained on 14 non-temporal tasks, evaluating both on fine-grained motion tasks that inherently require temporal understanding (Fig.~\ref{fig:fig_compaison_with_without_temp}).

The temporal-aware model consistently outperforms the variant without temporal tasks across all fine-grained evaluations, with gains most pronounced in tasks requiring precise temporal control: fine-grained generation ((T+DT)2M, R-Top3: +3.92) and detailed captioning (M2DT, Bleu@4: +4.68). Improvements are smaller for atomic edits. 
This pattern indicates that temporal-aware pre-training is fundamental, especially for the complex instructions that with multiple temporal information, and that the learned temporal priors help complex, multi-step scenarios like fine-grained editing. The results validate that explicitly modeling time is critical for capturing motion dynamics.

\subsection{Importance of Chain-of-Thought Annotations}
\label{sec:ablation_cot}
\begin{figure}[!t]
    \centering
    \includegraphics[width=0.8\linewidth]{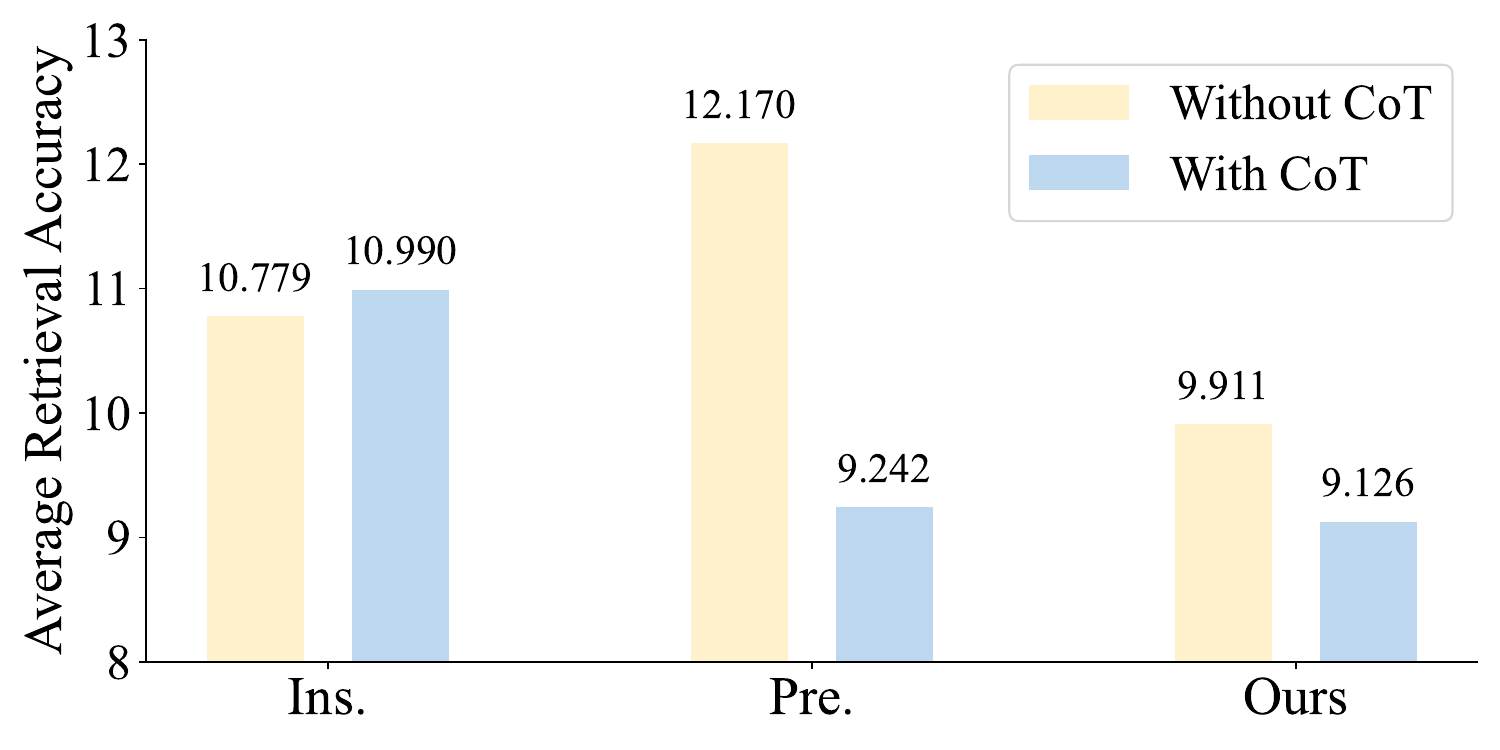}
    \caption{
    \textbf{Impact of chain-of-thought on complex editing.} Lower values are better. \textit{Ins.}: directly instruction-tuned. \textit{Pre.}: RAGS pre-trained. 
    }
    \label{fig:complex_editing_CoT_vs_woCoT}
\end{figure}

Fig.~\ref{fig:complex_editing_CoT_vs_woCoT} evaluates the impact of CoT annotations on complex motion editing, comparing performance with and without step-by-step reasoning prompts across training stages. 
The results reveal three interconnected insights. First, pretraining enables direct complex editing: the pretrained model achieves a performance of 12.170 without CoT, demonstrating that foundational motion-language knowledge supports end-to-end handling of complex edits. Second, CoT’s benefit is highly context-dependent: it significantly aids the pretrained model (AvgR: –2.928, lower is better), yet marginally impairs the instruction-tuned model (+0.211), indicating that explicit reasoning enhances complex task solving only when strong temporal and motion priors are present. Third, our full training scheme balances both, maintaining high performance and showing that RAGS pre-training provides robust priors that make explicit CoT prompting more effective rather than redundant.

\section{Conclusion and Future Work}
Current motion–language models are effective for holistic tasks but lack precise, spatio-temporally localized control required for fine-grained applications. This granularity gap arises from inadequate models for fine-grained grounding and the scarcity of large-scale fine-grained supervision. 
To bridge this gap, we present MotionMERGE with three synergistic advances: 
1)~a unified multi-granular motion-language model to address diverse motion generation, understanding, and editing tasks, demonstrating emergent zero-shot reasoning; 
2)~the Reasoning-Aware Granularity-Synergy pre-training strategy, which instills motion priors, temporal priors, localized alignment, and structured reasoning by efficiently aligning language with motion across granularities; 
and 3)~the MotionFineEdit dataset, a large-scale resource providing fine-grained, part-level, temporally-localized corrective annotations and motion-grounded chain-of-thought data, establishing a challenging new benchmark. 
Extensive experiments confirm that these advances collectively enable a leap in fine-grained, language-guided motion control, interpretability, and reasoning. 

Future research includes developing dedicated metrics for procedural correctness and enhancing the model’s explicit reasoning capability through supervision of intermediate reasoning steps.




\bibliographystyle{IEEEtran}
\bibliography{ref}


\begin{IEEEbiography}[{\includegraphics[width=1in,height=1.25in,clip,keepaspectratio]{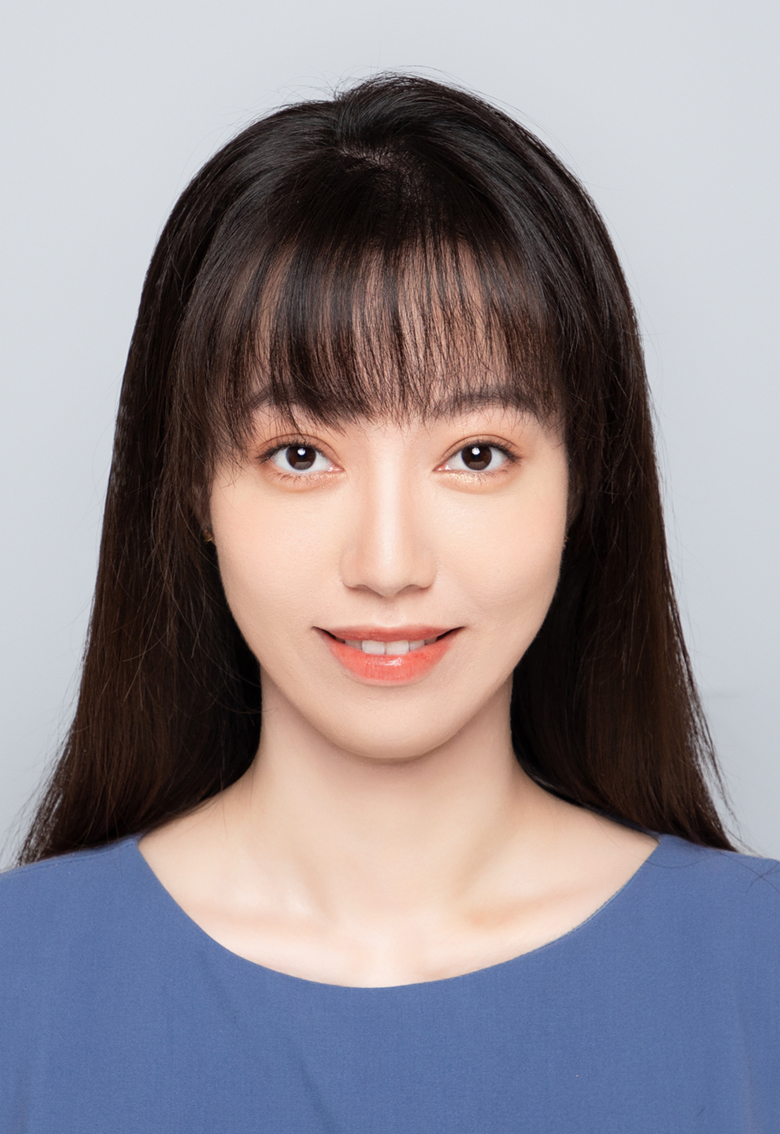}}]{Bizhu Wu} is a Ph.D. candidate in a collaborative training program jointly offered by the University of Nottingham Ningbo China, Shenzhen University, and University of Nottingham Nottingham UK. She is supervised by Dr. Jianfeng Ren, Linlin Shen, Ruibin Bai, and Rong Qu. Her research interests include effective representation of 3D human activity, 3D human motion generation, and motion understanding.
\end{IEEEbiography}

\begin{IEEEbiography}[{\includegraphics[width=1in,height=1.25in,clip,keepaspectratio]{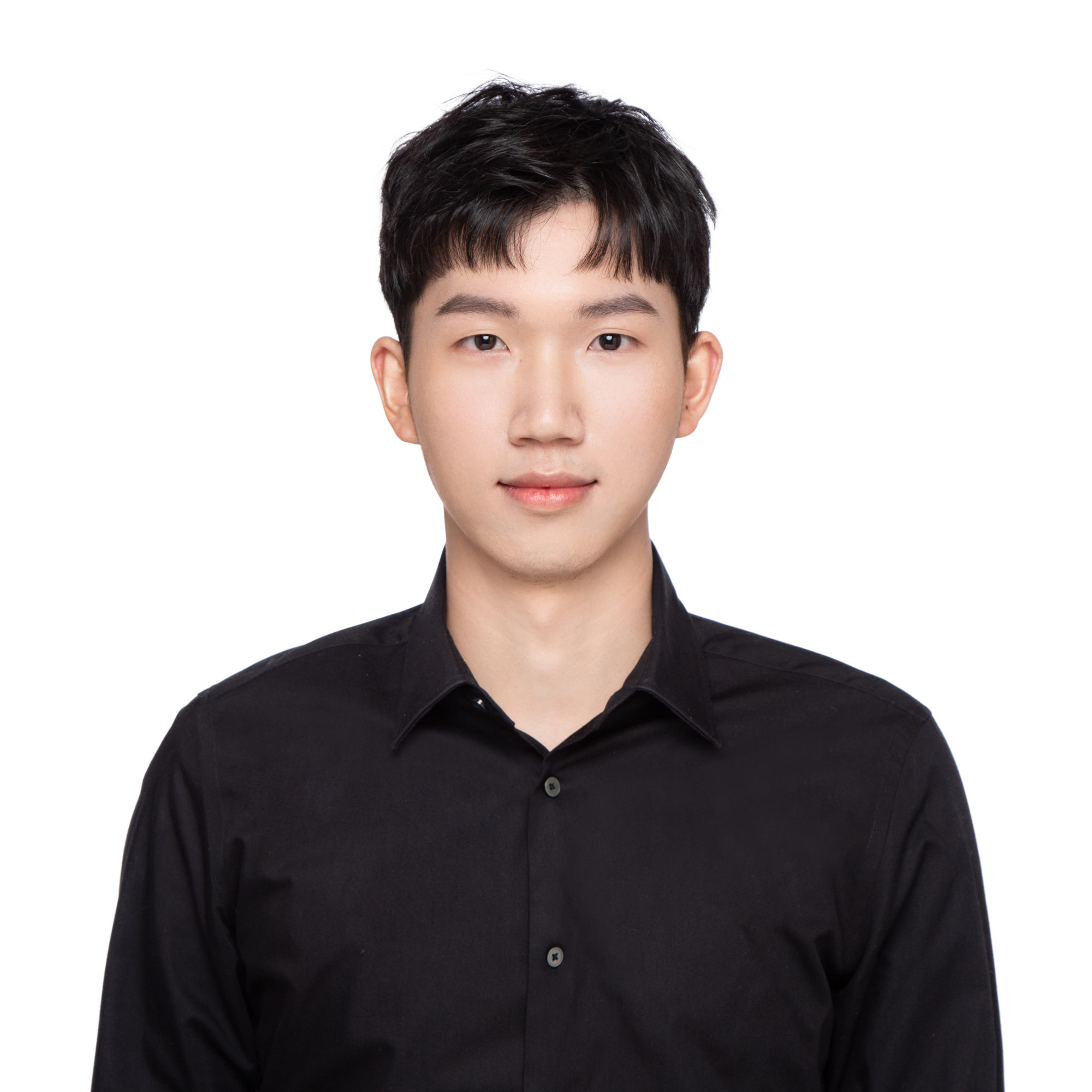}}]{Jinheng Xie} is a PhD student at Show Lab, National University of Singapore. His research focuses on unifying multimodal understanding and generation within a native unified multimodal model, with representative works including Show-o and Show-o2. His research has garnered over 2,500 citations. For more information, please visit https://sierkinhane.github.io/.
\end{IEEEbiography}

\begin{IEEEbiography}[{\includegraphics[width=1in,height=1.25in,clip,keepaspectratio]{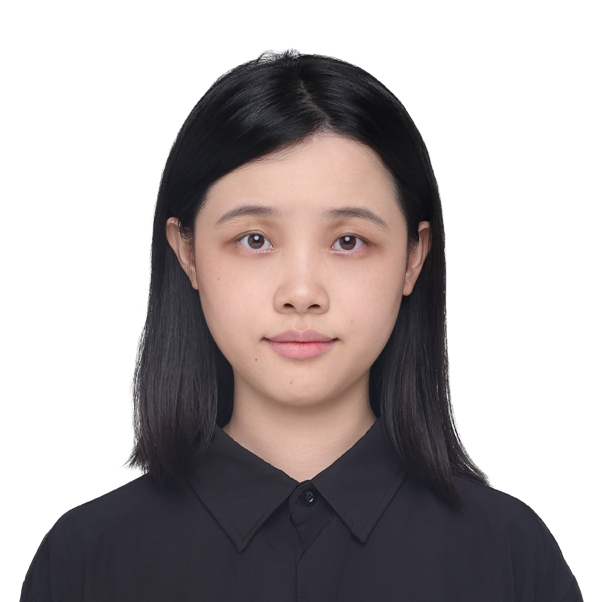}}]{Wenting Chen} is a Postdoctoral Fellow in the Department of Radiation Oncology at Stanford University, working with Professor Lei Xing. She received her Ph.D. degree from the Department of Electrical Engineering at City University of Hong Kong, supervised by Professor Yixuan Yuan. She was also a visiting researcher at Massachusetts General Hospital and Harvard Medical School, collaborating with Professor Xiang Li and Professor Quanzheng Li. Her research interests include vision-language models, generative AI, computer vision and their applications in medical AI.
\end{IEEEbiography}

\begin{IEEEbiography}[{\includegraphics[width=1in,height=1.25in,clip,keepaspectratio]{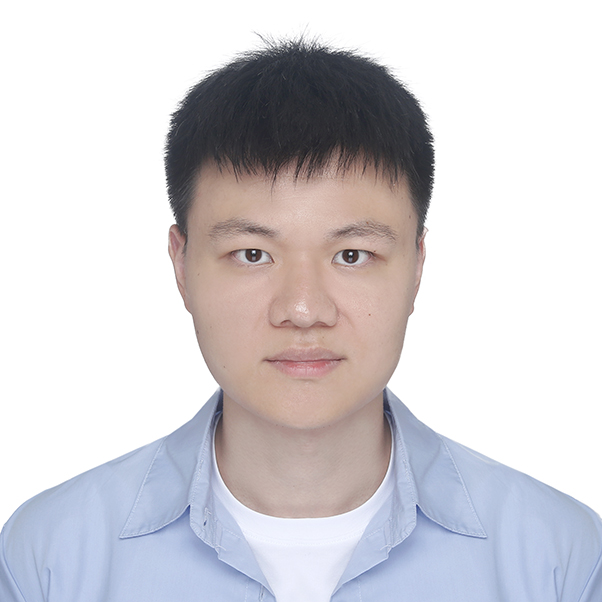}}]{Zhe Kong} is currently a Ph.D. candidate at Sun Yat‑sen University. His research interests center on image and video generation, with representative works such as MultiTalk and OMG. He has published papers in top‑tier conferences and journals, including NeurIPS, ECCV, and SIGGRAPH. 
\end{IEEEbiography}

\begin{IEEEbiography}[{\includegraphics[width=1in,height=1.25in,clip,keepaspectratio]{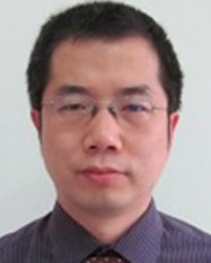}}]{Jianfeng Ren} (Senior Member, IEEE) received the B.Eng. degree from the National University of Singapore, Singapore, and the M.Sc. and Ph.D. degrees from Nanyang Technological University, Singapore. In 2018, he joined the School of Computer Science, University of Nottingham Ningbo China, where he is now a tenured Associate Professor. He has authored over 100 research papers, including contributions to TIP (3), TIFS (2), TMM (3) and PR (9), and conferences such as CVPR (3), ICCV (1), NeurIPS (1), AAAI (7), ACM MM (3), and IJCAI (1). His research interests include image/video processing, statistical pattern recognition, machine learning, and radar target recognition. 
He serves as an Associate Editor for Visual Computerand IET Biometrics, and as a Guest Editor for Symmetry. He has been a reviewer for more than 20 journals, including TIP and TSMC. He has also served as a TPC Member for CVPR, ICML, NeurIPS, AAAI, IJCAI and ECAI; Area Chair for ACM MM 2024/2025; and Program Chair for ICVRT 2024. 
\end{IEEEbiography}

\begin{IEEEbiography}[{\includegraphics[width=1in,height=1.25in,clip,keepaspectratio]{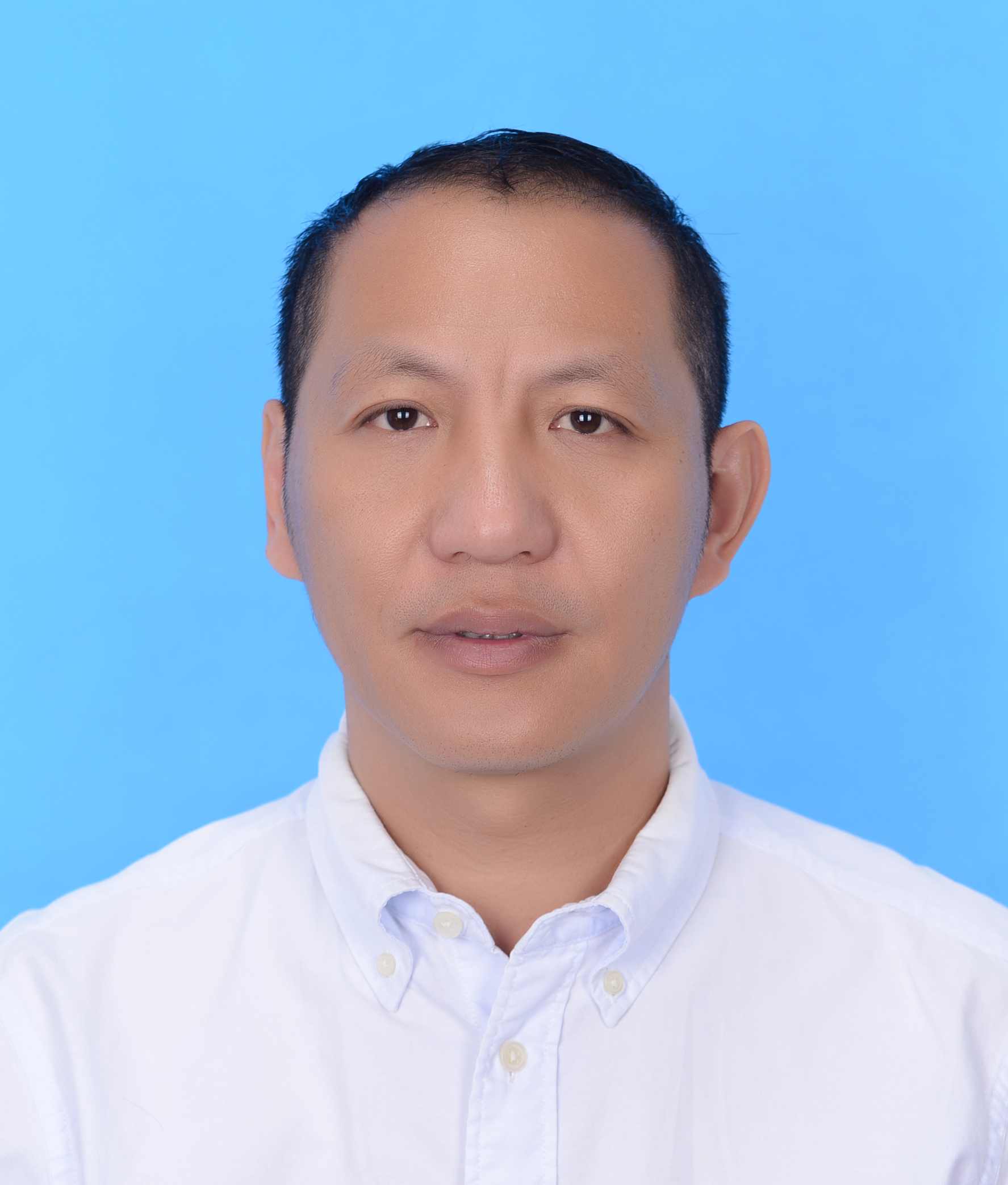}}]{Linlin Shen} (Senior Member) is currently a Pengcheng Scholar Distinguished Professor at School of Artificial Intelligence, Shenzhen University, Shenzhen, China. He is also a Honorary professor at School of Computer Science, University of Nottingham, UK. He serves as the Deputy director of National Engineering Lab of Big Data Computing Technology, Director of Computer Vision Institute, AI Research Center for Medical Image Analysis and Diagnosis and China-UK joint research lab for visual information processing. He also serves as the Co-Editor-in-Chief of the IET journal of Cognitive Computation and Systems, Senoir Area Editor of IEEE Trans. on Image Processing, Senior Editor of Expert Systems With Applications, and Associate Editor of Pattern Recognition and Scieitific Data. His research interests include deep learning, facial recognition, analysis/synthesis and medical image processing. Prof. Shen is listed as the ``Most Cited Chinese Researchers'' by Elsevier, ``Top 0.05\% Highly Ranked Scholar'' by ScholarGPS, and listed in a ranking of the ``Top 2\% Scientists in the World'' by Stanford University. He received the ``Best Paper Runner-up Award'' from the journal of IEEE Transactions on Affective Computing, ``Top Cited Article'' from Wiley, and ``Most Cited Paper Award'' from the journal of Image and Vision Computing. His cell classification algorithms were the winners of the International Contest on Pattern Recognition Techniques for Indirect Immunofluorescence Images held by ICIP and ICPR. His team has also been the runner-up and second runner-up of a number of competitions for object detection in remote sensing images, nucleus detection in histopathology images and facial expression recognition.
\end{IEEEbiography}

\begin{IEEEbiography}[{\includegraphics[width=1in,height=1.25in,clip,keepaspectratio]{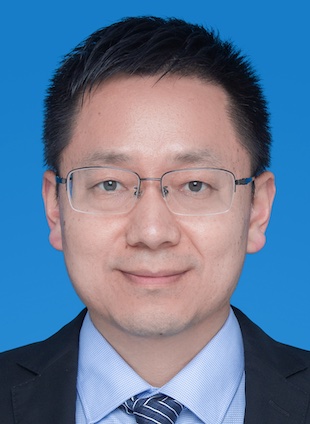}}]{Ruibin Bai} (Senior Member, IEEE) leads the Artificial Intelligence and Optimisation (AIOP) Research Centre at the University of Nottingham Ningbo China and the Key Lab of Digital Port Technologies. He currently serves as Associate Editor for IEEE Transactions on Evolutionary Computation, European Journal of Operations Research and Networks. His research interests include reinforcement learning, computational intelligence, integer programming and combinatorial optimization. His work is recognized in industry through “Huawei Spark Award" in 2022. Algorithms and software systems developed by his team have been commercialized in Ningbo Port and PingAn Good Doctor.
\end{IEEEbiography}

\begin{IEEEbiography}[{\includegraphics[width=1in,height=1.25in,clip,keepaspectratio]{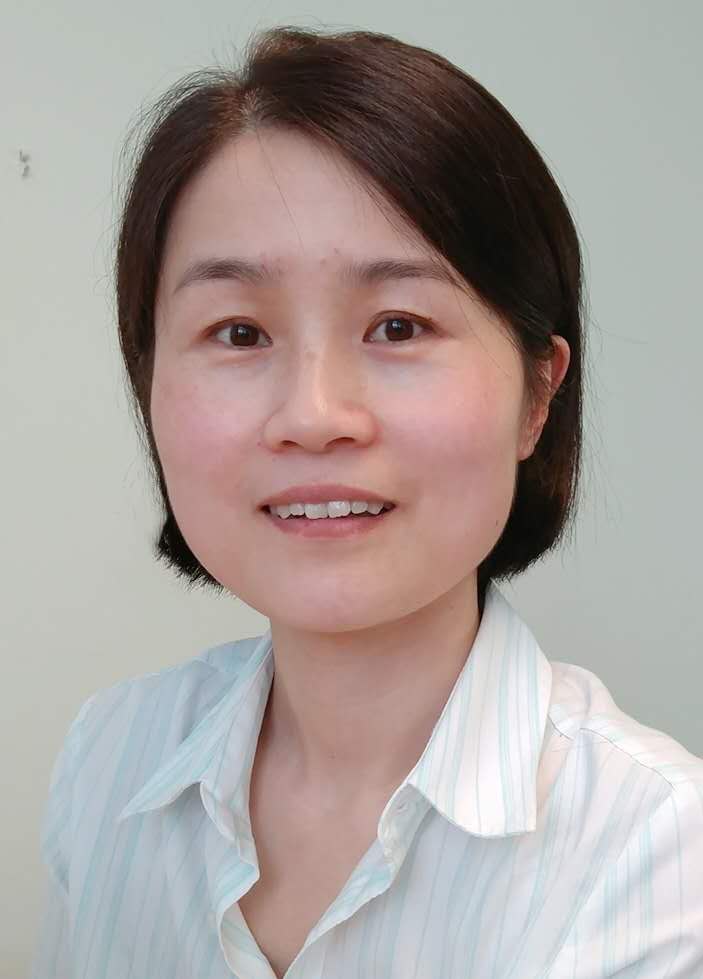}}]{Rong Qu} (Fellow, IEEE) is a Professor with the University of Nottingham, U.K. Her main research interests include the modeling and automated design of evolutionary algorithms for combinatorial optimization using hyper-heuristics, machine learning, constraint programming, and knowledge-based systems. Prof. Qu is among the World’s Top 2\% Scientists 2020–2025 by Stanford University and Elsevier. She has been Associate Editor and Guest Editor of special issues at IEEE Transaction on Evolutionary Computation, IEEE Computational Intelligence Magazine, etc. and the Chair/Vice-Chair of several task committees and task forces at IEEE CIS.
\end{IEEEbiography}


\clearpage
\setcounter{page}{1}

\appendices

\section{More Details on MotionFineEdit}

\subsection{Training of Text-driven Fine-Grained Motion Generator}
\label{sec:fg_mogen_sm}

The text-driven fine-grained human motion generator $\mathcal{G}$ is employed to synthesize the source motion $\bm{M}_\text{src}$ in Step 1 and the target motion $\bm{M}_\text{tgt}$ in Step 3. 
This generator is built upon the Granularity-Synergy pre-trained base model from MG-MotionLLM~\cite{mgmotionllm}, which learns motion-text alignment across multiple levels of textual granularity.
To further enhance the model’s capability to associate fine-grained body part descriptions with short motion segments, we fine-tune the model using paired data
$(L_\text{c}, \bm{L}_\text{d}^{\text{aug}}, \bm{M}_{\text{aug}})$.
Here, $\bm{M}_{\text{aug}}$ denotes a temporally cropped motion sequence composed of $k$ consecutive motion snippets, each containing 10 frames, resulting in a total length of $10k$ frames.
The snippet length is defined by FineMotion~\cite{finemotion}, and detailed textual descriptions are provided at this temporal resolution.
$L_\text{c}$ is the coarse-grained textual description from HumanML3D~\cite{humanml3d} corresponding to the cropped motion $\bm{M}_{\text{aug}}$.
$\bm{L}_\text{d}^\text{aug} = [L_{\text{d}_1}, L_{\text{d}_2}, \dots, L_{\text{d}_k}]$ contains fine-grained textual descriptions from FineMotion for the same motion, where each $L_{\text{d}_i}$ specifies body part movements within the $i$-th motion snippet.

During fine-tuning, the sentences within each $L_{\text{d}_i}$ are randomly shuffled.
This augmentation encourages the model to learn invariance to the intra-snippet sentence order, while preserving the temporal order across snippets.
Combined with the temporally augmented paired data, these strategies enhance the model’s ability to align motion snippets with body part movement descriptions.
As a result, the model gains fine-grained control over specific body parts through textual input, laying a solid foundation for constructing fine-grained motion editing data.

\subsection{Basic Fine-Grained Corrective Text Sampling for All Atomic Editing Operations}
\label{sec:fg_correctivetext_gen_SM}

In total, we define 11 atomic editing operations in our dataset, including 9 temporal and 2 spatial ones.
This design follows a factorized view of motion editing along two orthogonal factors, \ie, editing scope and editing type.
For temporal editing, the scope corresponds to three canonical positions (start, middle, and end), combined with three primitive types (padding, repeating, or deleting), yielding 9 temporal operations.
For spatial editing, the scope corresponds to local body parts within a given temporal interval, with two primitive types, \ie, adding and deleting.
We categorize these operations into five groups, as detailed below.

Notably, all complex editing instructions can be formulated as compositions of these atomic operations.
While not exhaustive, this set provides a simple and sufficiently expressive basis for the editing scenarios considered in this work.
Our dataset is therefore built upon these operations, categorized into five groups.
Since the operations share a similar construction mechanism, we present two below, 
while the others are detailed in the Appendix.

\vspace{0.5em}
\begin{itemize}
    \item \textbf{Temporal Padding}. 
    These operations extend immobile segments at the start, middle, or end of a motion. 
    As described in~\cite{mgmotionllm}, the special token \textcolor{orange}{\textless Motionless\textgreater} indicates the absence of body part movement (\ie, immobility) in a snippet.
    Given $\bm{L}_\text{d}$ consisting fine-grained descriptions for $k$ motion snippets, we randomly select a snippet index $p \in \{1, 2, \dots, k\}$ and insert $n \in \{1, 2, \dots, 20-k\}$ consecutive \textcolor{orange}{\textless Motionless\textgreater} tokens before the description for snippet indexed by $p$, \ie, $L_{\text{d}_p}^\text{src}$, 
    to ensure the total number of snippets does not exceed 20, the limit used to train $\mathcal{G}$.
    The resulting modified fine-grained textual descriptions $\bm{L}_\text{d}^\text{tgt}$ is:
    \begin{align}
        \bm{L}_\text{d}^\text{tgt} = 
        [\, 
        & \left\{L_{\text{d}_i}^\text{src}\right\}_{i=1}^{p-1} ; 
        \underbrace{\textcolor{orange}{\textless \text{Motionless}\textgreater} ; \dots ;
        \textcolor{orange}{\textless \text{Motionless}\textgreater}}_{n~\text{times}} ; \nonumber \\
        & \left\{L_{\text{d}_i}^\text{src}\right\}_{i=p}^k \,].
    \end{align}
    With such $\bm{L}_\text{d}^\text{tgt}$, the generator $\mathcal{G}$ can produce the $\bm{M}_{\text{tgt}}$ containing $10n$ additional frames of immobility at the corresponding position.
    Meanwhile, the basic corrective instruction $L_{\text{edit}}^{\text{basic}}$ is constructed by filling the sampled $p$ and $n$ into predefined templates (Tab.~\ref{table:edit_text_template_SM}, rows 1-3).

    \vspace{0.5em}
    \item \textbf{Temporal Repeating}. 
    These operations duplicate a motion segment at the start, middle, or end of the motion.
    We sample a start snippet index $p \in \{1, 2, \dots, k\}$ and a repeated segment length $n \in \{1, 2, \dots, 20-k\}$, representing the number of consecutive snippets to repeat.
    We replicate the corresponding snippets descriptions and insert them right after the segment to be repeated in $\bm{L}_\text{d}$, yielding the modified fine-grained motion descriptions:
    \begin{align}
        \bm{L}_\text{d}^\text{tgt} = [
        \left\{L_{\text{d}_i}^\text{src}\right\}_{i=1}^{p-1} ; 
        \, & L_{\text{d}_{p}}^\text{src} ; L_{\text{d}_{p+1}}^\text{src} ; \dots ; L_{\text{d}_{p+n-1}}^\text{src} ; \nonumber \\
        & L_{\text{d}_{p}}^\text{src} ; L_{\text{d}_{p+1}}^\text{src} ; \dots ; L_{\text{d}_{p+n-1}}^\text{src} ; 
        \left\{L_{\text{d}_i}^\text{src}\right\}_{i=p+n}^k
        \,].
    \end{align}
    With it, $\bm{M}_{\text{tgt}}$ therefore contains an additional $10n$ frames from the repeated segment.
    $L_{\text{edit}}^{\text{basic}}$ is constructed from $p$ and $n$ values using templates (Tab.~\ref{table:edit_text_template_SM}, rows 1-3).

    \vspace{0.5em}
    \item \textbf{Temporal Deleting}. 
    These operations remove designated motion segments and can be viewed as the inverse of temporal padding and repeating.
    Accordingly, the previously generated $\bm{M}_\text{tgt}$ from those operations is treated as $\bm{M}_\text{src}$, and the original $\bm{M}_\text{src}$ becomes the target $\bm{M}_\text{tgt}$ here.
    $L_{\text{edit}}^{\text{basic}}$ is generated using templates (Tab.~\ref{table:edit_text_template_SM}, rows 7–9).

    \vspace{0.5em}
    \item \textbf{Spatial Adding}. 
    This operation adds additional body part movement(s) within a selected snippet indexed by $p \in \{1, 2, \dots, k\}$.
    We build a pool of body part movement sentences $\bm{D}_\text{bpm}$ from FineMotion~\cite{finemotion}.
    A sentence $s_\text{bpm} \in \bm{D}_\text{bpm}$ is sampled and insert it into the body part movement description of the chosen snippet $L_{\text{d}_p}^\text{src}$.
    Therefore, the modified fine-grained motion descriptions is:
    \begin{equation}
        \bm{L}_\text{d}^\text{tgt} = [L_{\text{d}_1}^\text{src} ; \dots ; L_{\text{d}_{p-1}}^\text{src} ; L'_{\text{d}_p} ; \dots ; L_{\text{d}_{k}}^\text{src} ],
    \end{equation}
    where
    \begin{equation}
        L'_{\text{d}_p} = [
        \left\{s_i\right\}_{i=1}^{j-1}, 
        s_\text{bpm}, 
        \left\{s_i\right\}_{i=j}^{N_\text{sen}}
        ].
    \end{equation}
    Here, $s_i$ denotes a sentence in the chosen snippet, $N_\text{sen}$ is the total number of sentences in the chosen snippet, and $j \in \{1, 2, \dots, N_\text{sen}\}$ is a randomly insertion index.
    Because sentences within a snippet are randomly shuffled during training (Sec.~\ref{sec:fg_mogen_sm}), the insertion position $j$ does not affect the semantics.
    This design substantially increases the likelihood that the synthesized motion strictly follows the edited fine-grained descriptions $L'_{\text{d}_p}$, leading to reliable spatial edits.
    $L_{\text{edit}}^{\text{basic}}$ is formed by filling $p$ and $s_\text{bpm}$ into the template (Tab.~\ref{table:edit_text_template_SM}, row 10).

    \vspace{0.5em}
    \item \textbf{Spatial Deleting}. 
    Likewise, this operation is the inverse of spatial adding.
    The motion generated after spatial adding $\bm{M}_\text{tgt}$ is treated as $\bm{M}_\text{src}$, and the original motion $\bm{M}_\text{src}$ becomes $\bm{M}_\text{tgt}$ here.
    For the $L_{\text{edit}}^{\text{basic}}$, only the body part mentioned in $s_\text{bpm}$ is used rather than the full sentence, which better matches the logic of human expression (template see Tab.~\ref{table:edit_text_template_SM}, rows 11).

\end{itemize}

\rowcolors{2}{white}{gray!15}
\renewcommand{\arraystretch}{1.0}

\begin{table}[!t]
\centering
\footnotesize
\caption{
    Basic Corrective instructions for all atomic editing operations.
    FPS is the frame rate, used to convert frame counts to time (seconds).
}
\begin{tabular}{m{0.1cm}m{2.5cm}m{5.0cm}}

\toprule[1pt]

\textbf{No.} & \textbf{Editing Operation} & \textbf{Basic Corrective Instruction $L_{\text{edit}}^{\text{basic}}$}  \\

\midrule[1pt]

1 & Temporal Padding at the start
& Stay still for [$10n/\text{FPS}$]s at the start of the motion. \\

2 & Temporal Padding at the middle
& stay still for [$10n/\text{FPS}$]s after [$10(p-1)/\text{FPS}$]s of the motion. \\

3 & Temporal Padding at the end
& Stay still for [$10n/\text{FPS}$]s at the end of the motion. \\

\midrule

4 & Temporal Repeating at the start
& Repeat the first [$10n/\text{FPS}$]s of motion at the start. \\

5 & Temporal Repeating at the middle
& Repeat the [$10(p-1)/\text{FPS}$]s-[$10(p+n-1)/\text{FPS}$]s of motion after [$10(p+n-1)/\text{FPS}$] of the motion. \\

6 & Temporal Repeating at the end
& Repeat the last [$10n/\text{FPS}$]s of motion at the end. \\

\midrule

7 & Temporal Deleting at the start
& Delete the first [$10n/\text{FPS}$]s of motion. \\

8 & Temporal Deleting at the middle
& Delete [$10(p-1)/\text{FPS}$]-[$10(p+n-1)/\text{FPS}$] of motion. \\

9 & Temporal Deleting at the end
& Delete the last [$10n/\text{FPS}$]s of motion. \\

\midrule

10 & Spatial Adding
& Add the body part movement: [$s_\text{bpm}$] in [$10(p-1)/\text{FPS}$]-[$10p/\text{FPS}$] of the motion. \\

11 & Spatial Deleting
& Delete the movement of [body part] in [$10(p-1)/\text{FPS}$]-[$10p/\text{FPS}$] of the motion. \\

\bottomrule[1pt]

\end{tabular}
\label{table:edit_text_template_SM}
\end{table}

\rowcolors{2}{}{}
\renewcommand{\arraystretch}{1.0}

\subsection{Automatic Filtering for All Editing Operations}
\label{sec:auto_filter_SM}

Additional criteria of all types of editing operation are displayed below.

\vspace{0.5em}
\begin{itemize}

    \item \textbf{Temporal Padding.} 
    Let $\mathcal{I}_{\text{pad}} = \{ p, \dots, p+n-1 \}$ and $\mathcal{I}_{\text{non-pad}} = \{ 1, 2, \dots, p-1 \} \cup \{ p+n, \dots, k+n \}$ denote padded and non-padded snippet indices, respectively. 
    For non-padded snippets with indices $i \in \mathcal{I}_{\text{non-pad}}$, we require semantic consistency with the source motion:
    \begin{equation}
        \begin{cases}
            \text{sim}\big(\mathcal{F}(\bm{M}_\text{tgt}^i),\ \mathcal{F}(\bm{M}_\text{src}^i)\big) \ge \tau_1, & i \leq p-1 \\ 
            \text{sim}\big(\mathcal{F}(\bm{M}_\text{tgt}^i),\ \mathcal{F}(\bm{M}_\text{src}^{i-n})\big) \ge \tau_1, & i \geq p+n
        \end{cases},
    \end{equation}
    where $\bm{M}^{i}$ denotes the $i$-th motion snippet.
    Padded snippets with indices $i \in \mathcal{I}_{\text{pad}}$ should remain approximately static. 
    We enforce small joint deviations relative to the first padding frame $t_\text{ref} = (p-1) N_\text{s} +1$:
    \begin{align}
        \left\| \mathcal{R}(\bm{M}_\text{tgt}(t)) - \mathcal{R}(\bm{M}_\text{tgt}(t_\text{ref})) \right\|_1 \le \sigma, \nonumber \\
        t \in \bigcup_{i \in \mathcal{I}_\text{pad}} \{(i-1) N_\text{s}+1, \dots, i N_\text{s}\},
    \end{align}
    where $\bm{M}_{\text{tgt}}(t)$ denotes the pose at frame $t$,
    $\mathcal{R}(\cdot)$ converts the pose representation into 3D joint positions, 
    and $N_\text{s}=10$ frames per snippet.

    \vspace{0.5em}
    \item \textbf{Temporal Repeating.} 
    Let $\mathcal{I}_{\text{rep}} = \{p+n, \dots, p+2n-1\}$ denote repeated snippet indices and $\mathcal{I}_{\text{non-rep}} = \{1, \dots, p+n-1\} \cup \{p+2n, \dots, k+n\}$ non-repeated ones. 
    We require non-repeated snippets to match their source counterparts:
    \begin{equation}
        \begin{cases}
            \text{sim}(\mathcal{F}(\bm{M}_\text{tgt}^i), \mathcal{F}(\bm{M}_\text{src}^i)) \ge \tau_1, & i \le p+n-1, \\ 
            \text{sim}(\mathcal{F}(\bm{M}_\text{tgt}^i), \mathcal{F}(\bm{M}_\text{src}^{i-n})) \ge \tau_1, & i \ge p+2n,
        \end{cases}
    \end{equation}
    while repeated snippets must match the corresponding source segment:
    \begin{equation}
        \text{sim}(\mathcal{F}(\bm{M}_\text{tgt}^i), \mathcal{F}(\bm{M}_\text{src}^{i-n})) \ge \tau_1, \quad i \in \mathcal{I}_{\text{rep}}.
    \end{equation}

    \vspace{0.5em}
    \item \textbf{Spatial Adding.} 
    Let $j_\text{bp} = \text{BP}(s_\text{bpm})$ denote the body part to be edited, where $\text{BP}(\cdot)$ extracts the relevant body part from the textual description $s_\text{bpm}$.
    Pre-edit snippets (indices $i < p$) must remain consistent with the source:
    \begin{equation}
        \text{sim}\big(\mathcal{F}(\bm{M}_\text{tgt}^i), \mathcal{F}(\bm{M}_\text{src}^i)\big) \ge \tau_1.
    \end{equation}
    For the edited snippet and all subsequent ones ($i \ge p$), modifications are restricted to $j_\text{bp}$.  
    We enforce this by constructing mixed snippets that combine the edited body part from $\bm{M}_\text{tgt}$ with unedited parts from $\bm{M}_\text{src}$, which should differ from the source:
    \begin{equation}
        \text{sim}\Big(\mathcal{F}(\bm{M}_\text{tgt}^i[j_\text{bp}] \cup \bm{M}_\text{src}^i[\neg j_\text{bp}]), \ \mathcal{F}(\bm{M}_\text{src}^i)\Big) < \tau_2,
    \end{equation}
    where $\bm{M}[\cdot]$ extracts the specified body part from the motion and $\neg$ indicates all other parts.
    Conversely, replacing the edited body part in $\bm{M}_\text{tgt}$ with the source part while keeping all other parts from $\bm{M}_\text{tgt}$ should preserve similarity to the source:
    \begin{equation}
        \text{sim}\Big(\mathcal{F}(\bm{M}_\text{src}^i[j_\text{bp}] \cup \bm{M}_\text{tgt}^i[\neg j_\text{bp}]), \ \mathcal{F}(\bm{M}_\text{src}^i)\Big) \ge \tau_1.
    \end{equation}
    These constraints ensure that only the designated body part is modified, while all other parts remain consistent with the source.
    
\end{itemize}

\subsection{More Details of Human Annotation}
\label{sec:human_anno_SM}

Although the automatic filtering procedure removes most incorrectly paired generated data, certain challenging cases may still remain, such as unrealistic body translation when in the air or incorrect descriptions of added body part movements. 
Therefore, human annotators are employed to further filter these cases and ensure high-quality text-driven fine-grained motion editing pairs.

\begin{figure}[!h]
    \centering
    \includegraphics[width=0.9\linewidth]{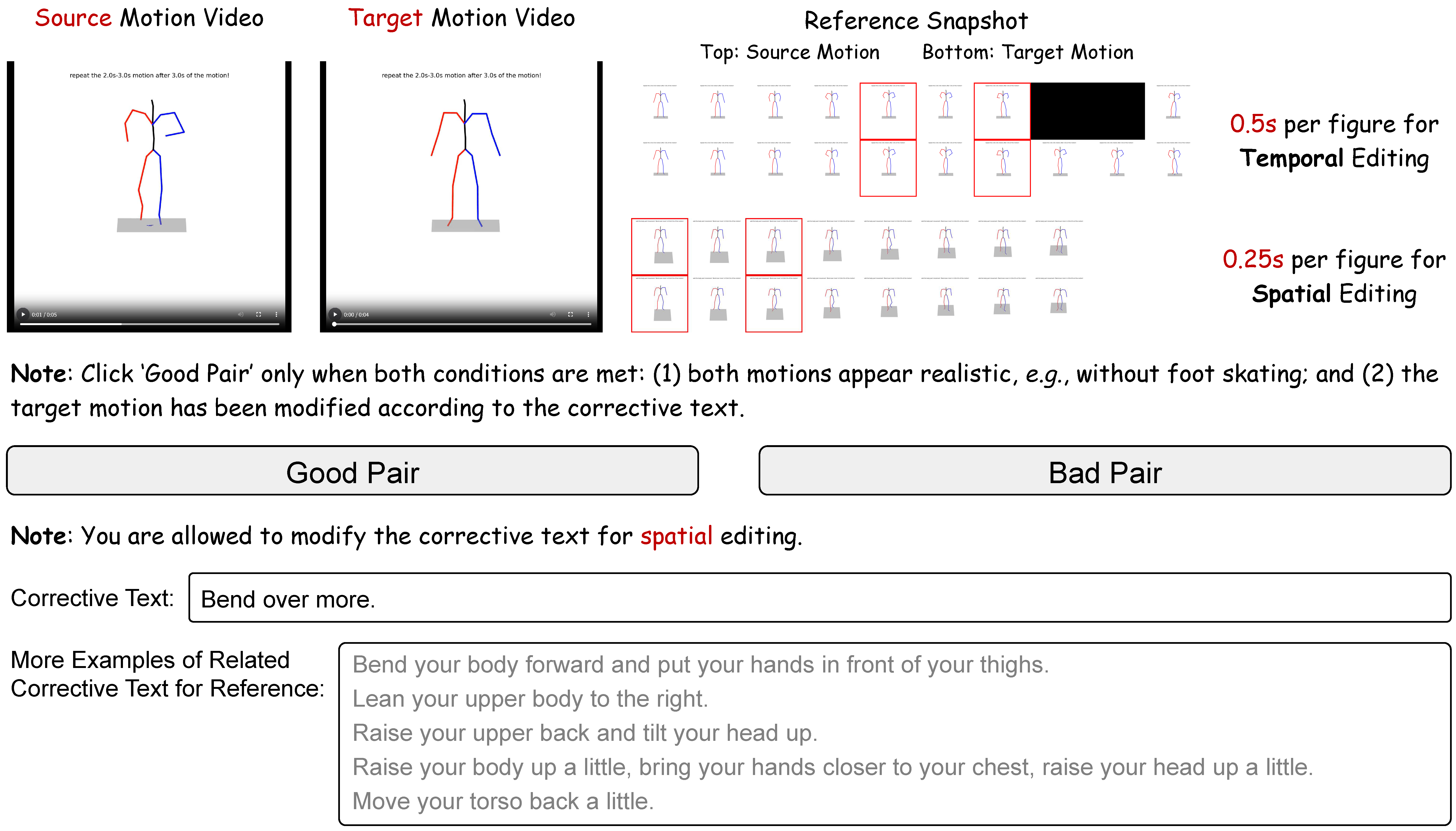}
    \caption{
        Our annotation platform.
    }
    \label{fig:annotation_platform}
\end{figure}

The annotation platform is illustrated in Fig.~\ref{fig:annotation_platform}. 
For each data entry, the source and target motion videos are displayed side by side, allowing annotators to directly compare motion realism and consistency. 
Video snapshots are also provided, with the start and end frames of the edited temporal interval explicitly highlighted to facilitate temporal comparison. 
Annotators are required to judge whether each pair is \emph{acceptable} (\ie, good pair) or \emph{unacceptable} (\ie, bad pair). 
A pair is considered acceptable only if both motions are realistic and the corrective text accurately describes the differences between them.

For temporal fine-grained motion editing pairs, we additionally present the snapshotted frames in a temporally aligned manner in which the source motion and the unedited portion of the target motion are aligned for more straightforward comparison.
For instance, as shown in Fig.~\ref{fig:annotation_platform}, the frames below the black blank frames, indicating temporal repetition, are newly added in the target motion. 
We align the frames after the edited interval in the source motion with the corresponding frames in the target motion.

For spatial fine-grained motion editing pairs, annotators are additionally required to verify that only the specified body part movements occur within the designated interval, the corrective text correctly describes the edit, and revise it when necessary. 
To guide revision, we provide example body part movement description sentences with the same body part, randomly sampled from the pool introduced in the previous subsection.

To ensure the quality of the final pairs, the human annotation pipeline consists of annotator training, annotation, and quality control. 
We recruit four annotators for this task. 
Before formal annotation, we train each annotator individually to ensure they understand the task and follow consistent criteria. 
They then iteratively annotate a small subsets of data, receive feedback, and repeat the process until achieving over 90\% annotation accuracy. 
Examples of unacceptable motions (\eg, unrealistic body translation when floating on the air) in~\cite{wang2024MotionCritic} are also provided as references. 
During annotation, data entries are randomly assigned to annotators.

Finally, we implement a strict quality-control procedure in which all annotated data are reviewed by an expert inspector. 
For temporal editing pairs, the annotation results are evaluated in batches of 100 entries. 
In each batch, 30\% of entries are randomly sampled and checked. 
A batch is accepted only if no more than three sampled entries disagree with the expert’s judgment; 
otherwise, the entire batch is re-annotated and the annotators are retrained. 
Moreover, all incorrectly annotated samples and their five neighboring entries are returned to the annotator for re-annotation. 
For spatial ones, each entry is individually checked and corrected by the expert.

\subsection{More Details on Enriching Fine-Grained Corrective Instructions}
\label{sec:llm_rewrite_SM}

We introduce an LLM-based linguistic augmentation stage to diversify corrective instructions. 
Instead of directly using the template-generated instructions $L_{\text{edit}}^{\text{basic}}$ and $L_{\text{complex\_edit}}^{\text{basic}}$, we rewrite them into more natural and diverse forms, producing enriched instructions $L_{\text{edit}}$ that better reflect realistic human expression patterns. 
Examples are shown in Fig.~\ref{fig:data_example}.

Specifically, we employ Gemini~\cite{team2023gemini} for this purpose and adopt different strategies for different types of editing instructions.
For atomic edits along the temporal dimension, their basic corrective text mainly conveys temporal information with relatively simple semantics, where variation lies primarily in phrasing rather than structure. 
We therefore construct a pool of 100 templates for each temporal operation and randomly sample them to generate diverse expressions.

In contrast, the basic corrective instructions for spatial atomic edits and complex edits contain richer semantics, including body parts and the order of steps. 
Their structures vary substantially across samples, making template-based rewriting insufficient to capture natural expression diversity. 
For these cases, we directly prompt Gemini to generate customized instructions for each sample, allowing the phrasing to adapt to the specific editing content. 
\begin{tcolorbox}[boxrule=0pt, colframe=white, sharp corners, left=1mm, right=1mm, top=1mm, bottom=1mm]
\begin{footnotesize}

    \textit{\textcolor{gray}{Spatial Atomic Edit}}
    
    You are a creative and considerate assistant. Provide all possible ways a human might express the sentence below. Respond only with valid JSON dict containing a single key–value pair, where the key is "rewritten\_instructions" and the value is a Python list in which each item represents one possible expression.
     
    [$L_{\text{edit}}^{\text{basic}}$]

    \vspace{2em}

    \textit{\textcolor{gray}{Complex Edit}}

    You are a creative and considerate assistant. The following items describe consecutive editing instructions that occur in chronological order. Do not change the order of the instructions. Please merge them into one coherent paragraph. Generate all possible natural ways a human might express the merged instruction. Respond only with a valid JSON dictionary containing a single key–value pair: The key must be "rewritten\_instructions". The value must be a Python list, where each item corresponds to one possible human expression. The editing instructions are:
    
    [$L_{\text{complex\_edit}}^{\text{basic}}$]

\end{footnotesize}
\end{tcolorbox}

\subsection{More Dataset Statistics Analysis}
\label{sec:statistics_sm}
Tab.~\ref{table:detail_split} details the number of triplets for each atomic editing operation and complex editing in each split of our dataset.

\begin{table*}[!h]
\begin{center}
    \caption{
        \textbf{Detailed numbers of each editing operation across all splits in our MotionFineEdit dataset.}
        \textbf{S} and \textbf{T} denote edits along the spatial and temporal dimensions, respectively.
        For each cell, the numbers before and after '\textbf{/}' refer to the counts of motion pairs and editing instructions.
    }
\label{table:detail_split}
\vspace{-0.5em}
\setlength{\tabcolsep}{2.5pt} 
\begin{tabular}{c|cccccc|cccc}
    \toprule[1pt]
      \multirow{2}{*}{\textbf{Split}} & \multicolumn{5}{c}{\textbf{Atomic Editing}}             
      & \multirow{2}{*}{\textbf{Total}}  
      & \multicolumn{3}{c}{\textbf{Complex Editing}} 
      & \multirow{2}{*}{\textbf{Total}} \\ 
      
      \cmidrule(r){2-6}\cmidrule(r){8-10} 
      
      & T-Padding & T-Repeating & T-Deleting & S-Adding & S-Deleting &
      & S-Only & S\&T & T-Only \\
      
      \midrule[0.5pt]
      
    Train & 2182 / 217154 & 1187 / 118653 & 3369 / 336792 & 648 / 9716 & 648 / 12226 & 8034 / 694541 & 1219 / 6691 & 4055 / 22672 & 6332 / 37678 & 11606 / 67041 \\
    
    Val & 128 / 12734 & 64 / 6390 & 192 / 19177 & 51 / 782 & 51 / 984 & 486 / 40067 & 72 / 383 & 152 / 867 & 128 / 784 & 352 / 2034 \\
    
    Test & 303 / 30135 & 192 / 19212 & 495 / 49501 & 111 / 1670 & 111 / 2098 & 1212 / 102616 & 227 / 1216 & 3196 / 17280 & 9890 / 56367 & 13313 / 74863 \\ 
    
    \midrule[0.5pt]
    
    \textbf{Total} & 2613 / 260023 & 1443 / 144255 & 4056 / 405470 & 810 / 12168 & 810 / 15308 & 9732 / 837224 & 1518 / 8290 & 7403 / 40819 & 16350 / 94829  & 25271 / 143938 \\ 
    
    \bottomrule[1pt]
    \end{tabular}
\end{center}
\vspace{-1em}
\end{table*}

All \textbf{motions} are generated by the text-driven fine-grained motion generator $\mathcal{G}$ trained on HumanML3D, whose motions cover diverse activities such as walking, swimming, and dancing. 
They are generated at 20 FPS, retargeted to a standard skeletal template, oriented to face the positive Z-axis, and provided in three representations:
(1) rotation parameters as in AMASS, 
(2) the pose representation of Guo \etal~\cite{humanml3d}, and 
(3) 22-keypoint coordinate trajectories over time.

\begin{figure}[!t]
\begin{center}
\includegraphics[width=1.0\linewidth]{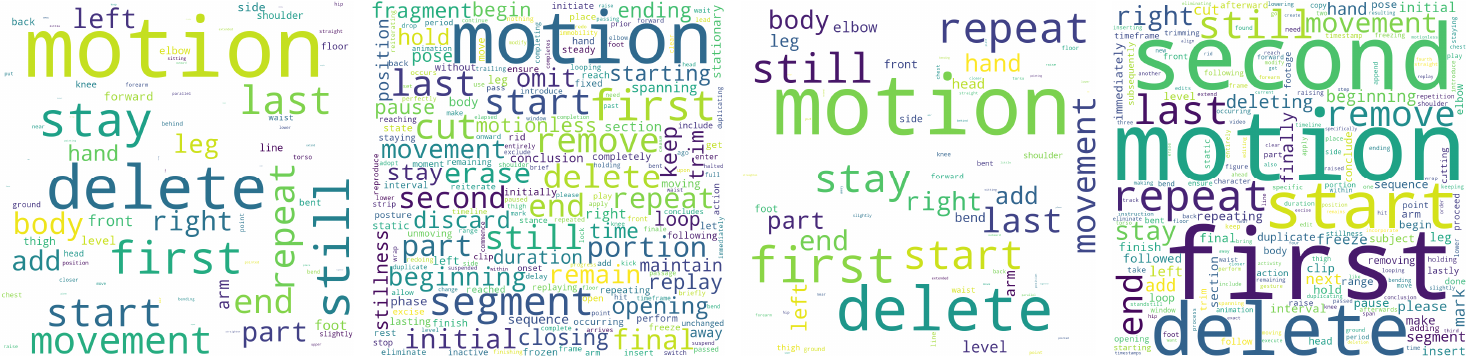}
\end{center}
\vspace{-1.0em}
   \caption{
        \textbf{Visualizations of the 200 most frequent words in our textual descriptions.} 
        From left to right, the figure presents the word clouds for basic corrective instructions and their rewritten counterparts for the \textit{atomic} editing, followed by those for the \textit{complex} editing.
   }
    \label{fig:wordcloud}
\vspace{-1em}
\end{figure}

Tab.~\ref{tab:statistics_text} presents statistics for the corrective \textbf{text} descriptions. 
Basic editing instructions have a limited vocabulary (135 and 128 unique words), due to the limited body parts and fixed description patterns in FineMotion~\cite{finemotion}, whereas using Gemini expands this to 762 and 978 words, as also illustrated by the word clouds in Fig.~\ref{fig:wordcloud}. 
For atomic edits, the average text length is 8.8 words, balancing conciseness and expressiveness. 
In contrast, complex edits combines multiple steps, leading to longer text descriptions with higher average and median lengths. Because the number of steps varies across complex edits, the standard deviation of words per text is also considerably larger.

\begin{table}[!t]
\centering
\caption{
    Statistics of the MotionFineEdit textual data.
}
\label{tab:statistics_text}
\vspace{-0.5em}
\setlength{\tabcolsep}{2pt} 
\begin{tabular}{lcccccc}

\toprule[1pt]

 & \multicolumn{3}{c}{\textbf{Atomic Editing}} & \multicolumn{3}{c}{\textbf{Complex Editing}} \\
 
\cmidrule(r){2-4}\cmidrule(r){5-7}

 & basic & rewritten & all & basic & rewritten & all \\

\midrule[0.5pt]
Total \#texts & 9732 & 827492 & 837224 & 25271 & 118,667 & 143938 \\
\#Unique words & 135 & 762 & 764 & 128 & 978 & 979 \\
Avg \#words per text & 8.7 & 9.2 & 8.8 & 45.0 & 46.9 & 46.6 \\
Std \#words per text & 4.6 & 2.1 & 2.2 & 12.6 & 13.4 & 13.3 \\
Median \#words per text & 9 & 8 & 8 & 45 & 47 & 46 \\
Min \#words per text & 5 & 4 & 4 & 14 & 6 & 6 \\
Max \#words per text & 43 & 42 & 43 & 110 & 98 & 110 \\
\bottomrule[1pt]
\end{tabular}
\vspace{-1em}
\end{table}

\vspace{1em}
\section{More Implementation Details}
\label{sec:more_im_details}
In addition to MotionMERGE with 220M parameters, we implement a smaller variant with 60M parameters and a larger variant with 770M parameters. 
Unless otherwise specified, most training settings follow those reported in Tab.~\ref{tab:appendix_im_details}.

\begin{table}[!h]
\caption{Hyperparameters for different variants of MotionMERGE.}
\vspace{-0.5em}
\label{tab:appendix_im_details}
\footnotesize
\centering
    \begin{tabular}{@{}lccc@{}}
        \toprule[1pt]
        MotionMERGE &  Small & Base & Large
        \\ 
        \midrule[0.5pt]
        Backbone & T5-Small & T5-Base & T5-Large \\
        Pre-training - Batch Size & 16 & 16 & 4 
        \\
        Model Size & 60M & 220M & 770M
        \\
        Pre-training - Iterations & 500K & 500K & 500K
        \\
        Pre-training - Learning Rate & 2e-4 & 2e-4 & 2e-4
        \\
        Instruction Tuning - Batch Size & 64 & 16 & 4 
        \\
        Instruction Tuning - Iterations & 300K & 300K & 300K
        \\
        Instruction Tuning - Learning Rate & 1e-4 & 1e-4 & 1e-4
        \\
        \bottomrule[1pt]
    \end{tabular}
\end{table}

We further apply task-specific hyperparameter adjustments as follows.
(1) Following \cite{guo2022tm2t}, motion captioning is generally easier than text-to-motion generation. 
To mitigate overfitting, the instruction-tuning iterations for motion captioning are reduced to 100K for all three model variants.
(2) For the motion detailed captioning task, the large model uses a batch size of 4 in both training stages. 
To account for the increased task difficulty under this limited batch size, we extend its instruction-tuning iterations to 400K, consistent with the setting used in MG-MotionLLM.
(3) For all motion editing tasks, including coarse-grained text-driven motion editing, fine-grained text-driven base motion editing, and fine-grained text-driven complex motion editing, the instruction-tuning iterations are set to 100K for all variants.

\vspace{1em}
\section{Qualitative Analysis of Complex Fine-Grained Text-Driven Motion Editing}

\begin{figure*}[!h]
\label{fig:complex_edit_cot_SM}
\begin{center}
\includegraphics[width=1.0\linewidth]{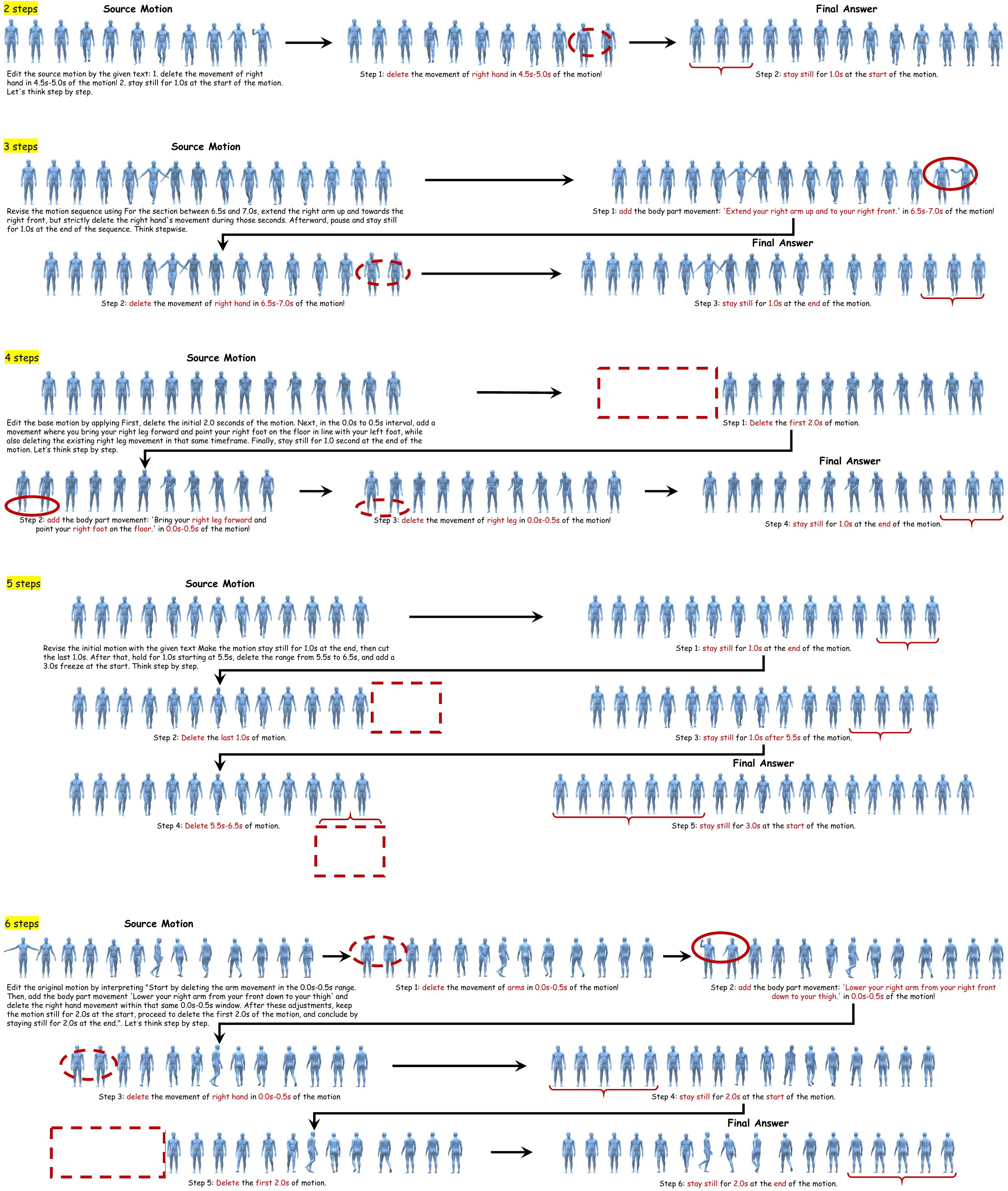}
\end{center}
\caption{
    \textbf{Qualitative reasoning processes and results for complex fine-grained text-driven human motion editing.}
    Dashed shapes (circles or rectangles) denote deletion operations in the spatial or temporal dimensions, while solid shapes (curly brackets or rectangles) indicate addition or repetition operations.
    The results show that MotionMERGE can precisely decompose complex fine-grained corrective instructions into atomic steps and generate the correct intermediate edited motions.
}
\label{fig:comlex_edit_with_cot}
\end{figure*}

Fig.~\ref{fig:complex_edit_cot_SM} presents qualitative examples of complex fine-grained text-driven human motion editing, where each example consists of multiple steps (ranging from 2 to 6), together with the corresponding reasoning processes.
As illustrated in the figure, complex corrective instructions are decomposed into a sequence of steps with atomic editing operations.
This step-by-step decomposition guides the model to produce intermediate edited motions that progressively follow the given instructions, thereby facilitating the synthesis of correct final edited results.

\onecolumn
\section{Complete Task List in Granularity-Synergy Pre-training Stage}
\label{sec:more_tasks}
In the Granularity-Synergy Pre-training stage, we pretrain the model with a total of \textbf{31} distinct motion-relevant tasks, 
including 13 existing classical coarse-grained tasks and 18 newly proposed fine-grained ones.
We summarize the information involved into three types: 
(1) textual descriptions (including both coarse captions and detailed texts), 
(2) temporal information, and 
(3) motion data.
All the tasks are divided into three groups according to the number of information types used in the input.
We display them separately in Tab.~\ref{tab:SM_tasks_one_input}, Tab.~\ref{tab:SM_tasks_two_input}, and Tab.~\ref{tab:SM_tasks_three_input}.

\footnotesize
\SetTblrStyle{caption-tag}{\small}
\DefTblrTemplate{caption-sep}{default}{.\enskip}

\begin{longtblr}
[
    caption={\small Examples of prompt templates for tasks that utilize \textbf{one} type of information in the input.},
    label={tab:SM_tasks_one_input},
]
{
    colspec = {
        X[1, l, m]   
        X[5, l, m]   
        X[1, l, m]   
    },
    width = \linewidth,
    rowhead = 1,
    row{even} = {bg=gray9},
    row{1} = {bg=white, font=\bfseries},
    stretch = -1,
    hline{1} = {1pt, solid},
    hline{2} = {0.5pt, solid},
    hline{9} = {1pt, solid},
}

\textbf{Task} & \textbf{Input} & \textbf{Output} \\


Text-to-Motion & 
\begin{minipage}{\linewidth} 
    \begin{itemize}
    \setlength{\itemsep}{0.3em}
        \item Show me a motion that conveys the meaning of [caption]. 
        \item Please create a motion that represents the power of [caption]. 
        \item Give me a gesture that corresponds to [caption].
    \end{itemize}
\end{minipage} 
& [motion] \\

Motion-to-Text & 
\begin{minipage}{\linewidth}
    \begin{itemize}
    \setlength{\itemsep}{0.3em}
        \item Describe the motion portrayed in [motion] using words. 
        \item Please describe the movement shown in [motion] using words. 
        \item What type of motion does [motion] depict?
    \end{itemize}
\end{minipage} 
& [caption] \\

Motion-to-(Text, Motion Script) & 
\begin{minipage}{\linewidth}
    \begin{itemize}
    \setlength{\itemsep}{0.3em}
        \item Explain the movement depicted in [motion] with the motion summary as well as the motion script. 
        \item Depict the movement in [motion] using the motion summary and the motion script. 
        \item Illustrate the action shown in [motion] using the motion summary and the motion script.
    \end{itemize}
\end{minipage} 
& 
\begin{minipage}{\linewidth}
    \#\#\# Motion Summary \#\#\# \\ \text{[caption]} \\ 
    \text{} \\  
    \#\#\# Motion Script \#\#\# \\ \text{[motion script]}
\end{minipage} 
\\

Motion-to-Motion Script &
\begin{minipage}{\linewidth}
    \begin{itemize}
    \setlength{\itemsep}{0.3em}
        \item Explain the movement depicted in [motion] with the motion script. 
        \item Illustrate the action shown in [motion] using the motion script. 
        \item What is the motion in [motion]? Describe it using the motion script.
    \end{itemize}
\end{minipage} 
& 
\begin{minipage}{\linewidth}
\#\#\# Motion Script \#\#\# \\ 
\text{[motion script]}
\end{minipage} 
\\

(Motion Script, Snippet Motion Script)-to-Time &
\begin{minipage}{\linewidth}
    \begin{itemize}
    \setlength{\itemsep}{0.6em}
        \item {Determine the start and end times of the snippet of the motion script within the whole motion script. \\ \#\#\# Whole Motion Script \#\#\# \\ \text{[motion script]} \\ \text{} \#\#\# Snippet Motion Script \#\#\# \\ \text{[snippet motion script]}}
        \item {Please outline the start and end points for the snippet of the motion script within the whole motion script. \\ \#\#\# Whole Motion Script \#\#\# \\ \text{[motion script]} \\ \text{} \#\#\# Snippet Motion Script \#\#\# \\ \text{[snippet motion script]}}
        \item {Could you detail the timing for the snippet of the motion script as it appears in the whole motion script? \\ \#\#\# Whole Motion Script \#\#\# \\ \text{[motion script]} \\ \text{} \#\#\# Snippet Motion Script \#\#\# \\ \text{[snippet motion script]}}
    \end{itemize}
\end{minipage} 
& [time] \\

(Motion, Snippet Motion)-to-Time & 
\begin{minipage}{\linewidth}
    \begin{itemize}
        \setlength{\itemsep}{0.3em}
        \item What are the time markers for [snippet motion] within [motion]?
        \item Provide the start and finish times of [snippet motion] within [motion].
        \item Outline the time span of [snippet motion] within the context of [motion].
    \end{itemize}
\end{minipage} 
& [time] \\

(Text, Motion Script)-to-Motion & 
\begin{minipage}{\linewidth}
    \begin{itemize}
    \setlength{\itemsep}{0.6em}
        \item {Please create a motion that represents the power of the motion summary and adheres to the motion script. \\ \#\#\# Motion Summary \#\#\# \\ \text{[caption]} \\ \text{} \#\#\# Motion Script \#\#\# \\ \text{[motion script]}}
        \item {Show me a motion that captures the essence of the motion summary and reflects the motion script. \\ \#\#\# Motion Summary \#\#\# \\ \text{[caption]} \\ \text{} \#\#\# Motion Script \#\#\# \\ \text{[motion script]}}
        \item {Design a motion that embodies the emotion of the motion summary and follows the motion script. \\ \#\#\# Motion Summary \#\#\# \\ \text{[caption]} \\ \text{} \#\#\# Motion Script \#\#\# \\ \text{[motion script]}}
    \end{itemize}
\end{minipage} 
& [motion] 

\end{longtblr}

\begin{longtblr}
[
    caption={\small Examples of prompt templates for tasks that utilize \textbf{two} types of information in the input.},
    label={tab:SM_tasks_two_input},
]
{
    colspec = {
        X[1, l, m]   
        X[5, l, m]   
        X[1, l, m]   
    },
    width = \linewidth,
    rowhead = 1,
    row{even} = {bg=gray9},
    row{1} = {bg=white, font=\bfseries},
    stretch = -1,
    hline{1} = {1pt, solid},
    hline{2} = {0.5pt, solid},
    hline{18} = {1pt, solid},
}

\textbf{Task} & \textbf{Input} & \textbf{Output} \\


(Time, Motion Script)-to-Snippet Motion Script & 
\begin{minipage}{\linewidth}
    \begin{itemize}
    \setlength{\itemsep}{0.6em}
        \item {What is [time]'s content in the whole motion script? \\ \#\#\# Whole Motion Script \#\#\# \\ \text{[motion script]}}
        \item {Detail [time] in the scope of the whole motion script. \\ \#\#\# Whole Motion Script \#\#\# \\ \text{[motion script]}}
        \item {Show the details of [time] within the whole motion script. \\ \#\#\# Whole Motion Script \#\#\# \\ \text{[motion script]}}
    \end{itemize} 
\end{minipage}
& {\#\#\# \text{[time]}'s Motion Script \#\#\# \\ \text{[snippet motion script]}}   \\

(Time, Motion)-to-Snippet Motion &
\begin{minipage}{\linewidth}
    \begin{itemize}
    \setlength{\itemsep}{0.3em}
        \item Illustrate the movement for [time] in the scope of [motion].
        \item Capture the motion for [time] within [motion].
        \item What does the movement of [time] look like in [motion]?
    \end{itemize} 
\end{minipage}
& [snippet motion] \\

(Motion, Snippet Motion Script)-to-Time &
\begin{minipage}{\linewidth}
    \begin{itemize}
    \setlength{\itemsep}{0.6em}
        \item {What are the start and end times of the snippet of the motion script in the [motion]? \\ \#\#\# Motion Script \#\#\# \\ \text{[snippet motion script]}}
        \item {Please outline the start and end points for the snippet of the motion script within the [motion]. \\ \#\#\# Motion Script \#\#\# \\ \text{[snippet motion script]}}
        \item {Could you detail the start and end times of the snippet of the motion script as found in the [motion]? \\ \#\#\# Motion Script \#\#\# \\ \text{[snippet motion script]}}
    \end{itemize} 
\end{minipage}
& [time] \\

(Motion Script, Snippet Motion)-to-Time &
\begin{minipage}{\linewidth}
    \begin{itemize}
    \setlength{\itemsep}{0.6em}
        \item {Can you pinpoint the duration of [snippet motion] within the motion guided by the motion script?  \\ \#\#\# Motion Script \#\#\# \\ \text{[motion script]}}
        \item {What are the beginning and end points of [snippet motion] in the motion following the motion script?    \\ \#\#\# Motion Script \#\#\# \\ \text{[motion script]}}
        \item {Can you indicate the start and stop times for [snippet motion] in the sequence aligned with the motion script?  \\ \#\#\# Motion Script \#\#\# \\ \text{[motion script]}}
    \end{itemize} 
\end{minipage}
& [time] \\

(Text, Head Motion)-to-Motion &
\begin{minipage}{\linewidth}
    \begin{itemize}
    \setlength{\itemsep}{0.3em}
        \item Create a gesture that starts with [head motion] and embodies [caption].  
        \item Start with [head motion] and generate a movement that captures [caption].
        \item Initiate a gesture with [head motion] that signifies [caption]. 
    \end{itemize} 
\end{minipage}
& [motion] \\

(Text, Tail Motion)-to-Motion & 
\begin{minipage}{\linewidth}
    \begin{itemize}
    \setlength{\itemsep}{0.3em}
        \item Create a motion that concludes with [tail motion] and represents [caption]. 
        \item Generate a gesture that conveys [caption] and concludes with [tail motion]. 
        \item Show a gesture that captures [caption] and finishes with [tail motion].
    \end{itemize}
\end{minipage}
& [motion] \\

(Text, Random Motions)-to-Motion &
\begin{minipage}{\linewidth}
    \begin{itemize}
    \setlength{\itemsep}{0.3em}
        \item Design a gesture using the tokens [random motions] to express [caption].  
        \item Craft a gesture reflecting [caption] through the tokens [random motions]. 
        \item Form a gesture reflecting [caption] with the tokens [random motions].
    \end{itemize}
\end{minipage}
& [motion] \\

(Text, Motion Script, Head Motion)-to-Motion &
\begin{minipage}{\linewidth}
    \begin{itemize}
    \setlength{\itemsep}{0.6em}
        \item {Starting with [head motion], create a motion that aligns with the motion summary and adheres to the motion script. \\ \#\#\# Motion Summary \#\#\# \\ \text{[caption]} \\ \text{} \#\#\# Motion Script \#\#\# \\ \text{[motion script]}}
        \item {Initiate a motion with [head motion] that matches the motion summary and follows the motion script. \\ \#\#\# Motion Summary \#\#\# \\ \text{[caption]} \\ \text{} \#\#\# Motion Script \#\#\# \\ \text{[motion script]}}
        \item {From the initial [head motion], develop a motion that complies with the motion summary and follows the motion script. \\ \#\#\# Motion Summary \#\#\# \\ \text{[caption]} \\ \text{} \#\#\# Motion Script \#\#\# \\ \text{[motion script]}}
    \end{itemize}
\end{minipage}
& [motion] \\

(Text, Motion Script, Tail Motion)-to-Motion &
\begin{minipage}{\linewidth}
    \begin{itemize}
    \setlength{\itemsep}{0.6em}
        \item {Create a motion that reflects the motion summary and adheres to the motion script, ending with [tail motion]. \\ \#\#\# Motion Summary \#\#\# \\ \text{[caption]} \\ \text{} \#\#\# Motion Script \#\#\# \\ \text{[motion script]}}
        \item {Produce a motion that embodies the motion summary and matches the motion script, concluding with [tail motion].  \\ \#\#\# Motion Summary \#\#\# \\ \text{[caption]} \\ \text{} \#\#\# Motion Script \#\#\# \\ \text{[motion script]}} 
        \item {Craft a motion that illustrates the motion summary, follows the motion script, and ends with [tail motion].  \\ \#\#\# Motion Summary \#\#\# \\ \text{[caption]} \\ \text{} \#\#\# Motion Script \#\#\# \\ \text{[motion script]}}
    \end{itemize}
\end{minipage}
& [motion] \\

(Text, Motion Script, Random Motions)-to-Motion &
\begin{minipage}{\linewidth}
    \begin{itemize}
    \setlength{\itemsep}{0.6em}
        \item {Construct a motion with the tokens [random motions] that matches the motion summary and adheres to the motion script. \\ \#\#\# Motion Summary \#\#\# \\ \text{[caption]} \\ \text{} \#\#\# Motion Script \#\#\# \\ \text{[motion script]}}
        \item {Design a movement with key tokens [random motions] that conveys the motion summary and follows the motion script.  \\ \#\#\# Motion Summary \#\#\# \\ \text{[caption]} \\ \text{} \#\#\# Motion Script \#\#\# \\ \text{[motion script]}}
        \item {Formulate a motion with the tokens [random motions] that conveys the motion summary and follows the motion script.  \\ \#\#\# Motion Summary \#\#\# \\ \text{[caption]} \\ \text{} \#\#\# Motion Script \#\#\# \\ \text{[motion script]}}
    \end{itemize}
\end{minipage}
& [motion] \\

(Text, Time)-to-Motion &
\begin{minipage}{\linewidth}
    \begin{itemize}
    \setlength{\itemsep}{0.3em}
        \item Can you generate a [time] segment of the movement that embodies [caption]?
        \item Give me [time] of the motion that reflects the meaning of [caption].  
        \item Can you produce a motion segment of [time] representing [caption]? 
    \end{itemize}
\end{minipage}
& [motion] \\

(Motion, Time)-to-Snippet Motion Script &
\begin{minipage}{\linewidth}
    \begin{itemize}
    \setlength{\itemsep}{0.3em}
        \item Detail the motion for [time] in [motion], using the motion script.
        \item Describe the movement for [time] in [motion], with the motion script. 
        \item Clarify the action for [time] in [motion] using the motion script.
    \end{itemize}
\end{minipage}
& { \#\#\# Motion Script \#\#\# \\ \text{[snippet motion script]}} \\

(Text, Snippet Motion)-to-Time &
\begin{minipage}{\linewidth}
    \begin{itemize}
    \setlength{\itemsep}{0.3em}
        \item Pinpoint the exact times for [snippet motion] within the motion sequence that captures [caption]. 
        \item Identify when the segment [snippet motion] starts and finishes in the motion expressing [caption]. 
        \item Can you identify the timing of the snippet [snippet motion] in the motion that symbolizes[caption]?
    \end{itemize}
\end{minipage}
& [time] \\

(Text, Motion Script, Snippet Motion)-to-Time &
\begin{minipage}{\linewidth}
    \begin{itemize}
    \setlength{\itemsep}{0.6em}
         \item {Identify the timing of the snippet [snippet motion] within the motion that reflects the motion summary and adheres to the motion script.   \\ \#\#\# Motion Summary \#\#\# \\ \text{[caption]} \\ \text{} \#\#\# Motion Script \#\#\# \\ \text{[motion script]}}
         \item {Detail the start and end times for the segment [snippet motion] in the motion that matches the motion summary and follows the motion script.  \\ \#\#\# Motion Summary \#\#\# \\ \text{[caption]} \\ \text{} \#\#\# Motion Script \#\#\# \\ \text{[motion script]}}
         \item {Could you detail the timing of the segment [snippet motion] in the gesture that complies with the motion summary and follows the motion script?  \\ \#\#\# Motion Summary \#\#\# \\ \text{[caption]} \\ \text{} \#\#\# Motion Script \#\#\# \\ \text{[motion script]}}
    \end{itemize}
\end{minipage}
& [time] \\

(Text, Motion Script, Time)-to-Motion &
\begin{minipage}{\linewidth}
    \begin{itemize}
    \setlength{\itemsep}{0.6em}
        \item {Can you generate a [time] segment of the movement that embodies the motion summary and the motion script? \\ \#\#\# Motion Summary \#\#\# \\ \text{[caption]} \\ \#\#\# Motion Script \#\#\# \\ \text{[motion script]}}
        
        \item {Give me [time] of the motion that reflects the meaning of the motion summary and the motion script.  \\ \#\#\# Motion Summary \#\#\# \\ \text{[caption]} \\ \#\#\# Motion Script \#\#\# \\ \text{[motion script]}}
        
        \item {Given the motion summary and the motion script, can you give me its [time] clip?  \\ \#\#\# Motion Summary \#\#\# \\ \text{[caption]} \\ \#\#\# Motion Script \#\#\# \\ \text{[motion script]}}
    \end{itemize}
\end{minipage}
& [motion] \\

(Source Motion, Coarse Edit Text)-to-Target Motion & 
\begin{minipage}{\linewidth}
    \begin{itemize}
    \setlength{\itemsep}{0.6em}
        \item {Change the source motion [source motion] in accordance with the given text: [coarse edit text]}
        
        \item {Edit the motion sequence [source motion] with the given text: [coarse edit text]}
        
        \item {Revise the original motion [source motion] using the given text: [coarse edit text]}
    \end{itemize}
\end{minipage}
& [target motion]

\end{longtblr}

\begin{longtblr}
[
    caption={\small Examples of prompt templates for tasks that utilize \textbf{three} types of information in the input.},
    label={tab:SM_tasks_three_input},
]
{
    colspec = {
        X[1, l, m]   
        X[5, l, m]   
        X[1, l, m]   
    },
    width = \linewidth,
    rowhead = 1,
    row{even} = {bg=gray9},
    row{1} = {bg=white, font=\bfseries},
    stretch = -1,
    hline{1} = {1pt, solid},
    hline{2} = {0.5pt, solid},
    hline{10} = {1pt, solid},
}

\textbf{Task} & \textbf{Input} & \textbf{Output} \\


(Text, Time, Head Motion)-to-Motion &
\begin{minipage}{\linewidth}
    \begin{itemize}
    \setlength{\itemsep}{0.3em}
        \item For a gesture that aligns with [caption], provide [time]'s motion snippet that begins with [head motion].
        \item Please produce a [time] motion segment, using [head motion] as the start point, from a gesture representing [caption]. 
        \item I need a [time] snippet with [head motion] as the initial input and from a gesture that symbolizes [caption].
    \end{itemize}
\end{minipage}
& [motion] \\

(Text, Time, Tail Motion)-to-Motion &
\begin{minipage}{\linewidth}
    \begin{itemize}
    \setlength{\itemsep}{0.3em}
        \item Create a [time] motion clip ending with [tail motion] from a gesture reflecting [caption].
        \item Generate a [time] motion clip that ends with [tail motion] from a gesture symbolizing [caption]. 
        \item Produce a [time] motion segment ending with [tail motion] from a gesture that represents [caption].
    \end{itemize}
\end{minipage}
& [motion] \\

(Text, Time, Random Motions)-to-Motion &
\begin{minipage}{\linewidth}
    \begin{itemize}
    \setlength{\itemsep}{0.3em}
        \item Create a [time] snippet featuring [random motions] from a motion that signifies [caption].
        \item Deliver a [time] excerpt including [random motions] from a gesture reflecting [caption]. 
        \item Create a [time] snippet with key tokens [random motions] from a gesture reflecting [caption]. 
    \end{itemize}
\end{minipage}
& [motion] \\

(Text, Motion Script, Time, Head Motion)-to-Motion &
\begin{minipage}{\linewidth}
    \begin{itemize}
    \setlength{\itemsep}{0.6em}
        \item {I need a [time]\ snippet, starting at [head motion]\, from the motion detailed in the motion summary and based on the motion summary.  \\ \#\#\# Motion Summary \#\#\# \\ \text{[caption]} \\ \#\#\# Motion Script \#\#\# \\ \text{[motion script]}}
        
        \item {Produce a [time]\ clip, starting with [head motion]\, taken from the motion defined by the motion summary and the motion summary.  \\ \#\#\# Motion Summary \#\#\# \\ \text{[caption]} \\ \#\#\# Motion Script \#\#\# \\ \text{[motion script]}}
        
        \item {Generate a [time]\ motion clip beginning with [head motion], sourced from the motion outlined in the motion summary and structured by the motion summary.  \\ \#\#\# Motion Summary \#\#\# \\ \text{[caption]} \\ \#\#\# Motion Script \#\#\# \\ \text{[motion script]}}
    \end{itemize}
\end{minipage}
& [motion] \\

(Text, Motion Script, Time, Tail Motion)-to-Motion &
\begin{minipage}{\linewidth}
    \begin{itemize}
    \setlength{\itemsep}{0.6em}
        \item {Provide a [time] clip that ends with [tail motion], derived from the full gesture that represents the motion summary and follows the motion script.  \\ \#\#\# Motion Summary \#\#\# \\ \text{[caption]} \\ \#\#\# Motion Script \#\#\# \\ \text{[motion script]}}
        
        \item {Produce a [time] snippet, concluding with [tail motion], from the motion that mirrors the motion summary and is built according to the motion script.  \\ \#\#\# Motion Summary \#\#\# \\ \text{[caption]} \\ \#\#\# Motion Script \#\#\# \\ \text{[motion script]}}
        
        \item {Please provide a [time] snippet that ends with [tail motion], from the entire motion described by the motion summary and follows the motion script.  \\ \#\#\# Motion Summary \#\#\# \\ \text{[caption]} \\ \#\#\# Motion Script \#\#\# \\ \text{[motion script]}}
    \end{itemize}
\end{minipage}
& [motion] \\

(Text, Motion Script, Time, Random Motions)-to-Motion &
\begin{minipage}{\linewidth}
    \begin{itemize}
    \setlength{\itemsep}{0.6em}
        \item {Create a [time] motion snippet with [random motions], based on a gesture that matches the motion summary and adheres to the motion script.  \\ \#\#\# Motion Summary \#\#\# \\ \text{[caption]} \\ \#\#\# Motion Script \#\#\# \\ \text{[motion script]}}
        
        \item {Create a [time] motion snippet featuring [random motions], based on a gesture that matches the motion summary and follows the motion script.   \\ \#\#\# Motion Summary \#\#\# \\ \text{[caption]} \\ \#\#\# Motion Script \#\#\# \\ \text{[motion script]}}
        
        \item {Generate a [time] motion clip with [random motions], originating from a gesture that matches the motion summary and follows the motion script.  \\ \#\#\# Motion Summary \#\#\# \\ \text{[caption]} \\ \#\#\# Motion Script \#\#\# \\ \text{[motion script]}}
    \end{itemize}
\end{minipage}
& [motion] \\

(Source Motion, Fine Atomic Edit Text, Time)-to-Target Motion &
\begin{minipage}{\linewidth}
    \begin{itemize}
    \setlength{\itemsep}{0.6em}
        \item {Modify the source motion [source motion] according to the provided text: [fine atomic edit text (with time)]}
        
        \item {Edit the motion sequence [source motion] with the given text: [fine atomic edit text (with time)]}
        
        \item {Revise the provided motion [source motion] using the given text: [fine atomic edit text (with time)]}
    \end{itemize}
\end{minipage}
& [target motion] \\

(Source Motion, Fine Complex Edit Text, Time)-to-(Reasoning Process, Target Motion) &
\begin{minipage}{\linewidth}
    \begin{itemize}
    \setlength{\itemsep}{0.6em}
        \item {Change the source motion [source motion] according to [fine complex edit text (with time)], reasoning step by step.}
        
        \item {Revise [source motion] using the provided description [fine complex edit text (with time)], step by step.}
        
        \item {Modify the motion [source motion] per the given text [fine complex edit text (with time)], step by step.}
    \end{itemize}
\end{minipage}
& [reasoning process with fine atomic edit text (with time) and intermediate motion in each step] The final answer is: [target motion]

\end{longtblr}

\vfill

\end{document}